\pdfoutput=1

\documentclass[11pt]{article}

\usepackage[final]{acl}

\usepackage{times}
\usepackage{latexsym}

\usepackage[T1]{fontenc}

\usepackage[utf8]{inputenc}

\usepackage{microtype}

\usepackage{inconsolata}

\usepackage{graphicx}

\usepackage{listings}
\usepackage{xcolor}
\usepackage{tcolorbox}
\usepackage{graphicx}
\usepackage{float}
\usepackage{amsmath}
\usepackage{booktabs}
\usepackage{colortbl}
\usepackage {makecell}
\usepackage{arydshln}
\usepackage{hyperref}
\usepackage{subfigure}
\definecolor{mygray}{gray}{.9}

\usepackage{amsfonts}
\usepackage{amsthm}   
\theoremstyle{plain}  
\newtheorem*{assumption}{Assumption}

\theoremstyle{definition}  

\title{Unveiling and Addressing Pseudo Forgetting in Large Language Models}

\author{
Huashan Sun\quad \textbf{Yizhe Yang} \quad \textbf{Yinghao Li} \quad \textbf{Jiawei Li}   
 \quad \textbf{Yang Gao}\thanks{~~Corresponding author} \\
School of Computer Science and Technology, Beijing Institute of Technology\\
\texttt{\{hssun,yizheyang,yhli,jwli,gyang\}@bit.edu.cn}
}

\begin{document}
\maketitle

\begin{abstract}
Although substantial efforts have been made to mitigate catastrophic forgetting in continual learning, the intrinsic mechanisms are not well understood. In this work, we demonstrate the existence of  "pseudo forgetting": the performance degradation on previous tasks is not attributed to a loss of capabilities, but rather to the failure of the instructions to activate the appropriate model abilities. We show that the model's performance on previous tasks can be restored through two simple interventions: (1) providing partial external correct rationale, and (2) appending semantically meaningless suffixes to the original instructions, to guide the generation of correct rationales. Through empirical analysis of the internal mechanisms governing rationale generation, we reveal that models exhibiting pseudo forgetting show reduced instruction dependence during rationale generation, leading to suboptimal activation of their inherent capabilities. Based on this insight, we propose Rationale-Guidance Difficulty based Replay (RGD-R) framework that dynamically allocates replay data based on the model’s ability to correctly leverage the intrinsic capabilities. Experimental results demonstrate that RGD-R effectively mitigates pseudo forgetting while maintaining model plasticity.
\end{abstract}

\section{Introduction}
Continual learning enables Large Language Models (LLMs)~\citep{i:1-GPT3,yang2023mindllm} to incrementally learn from a sequence of tasks, helping LLMs adapt to the dynamic nature of real-world data and improve their capabilities over time~\citep{Lifelong_Learning_Survey, li-etal-2024-fundamental}. However, LLMs still face catastrophic forgetting,  where performance on previous tasks deteriorates when learning new ones~\citep{MCCLOSKEY1989109}.


Despite the extensive methods proposed to mitigate catastrophic forgetting~\citep{InsCL, O-LoRA, SAPT}, limited studies investigate the intrinsic mechanisms underlying this phenomenon. \citet{Implicit_Inference} proposed the ``task inference'' hypothesis, which suggests that fine-tuning biases the model toward utilizing newly acquired capabilities, rather than causing a loss of previously learned abilities. While this hypothesis is validated on synthetic datasets and small transformers, direct empirical evidence from natural language datasets and LLMs is missing. 
Similarly, \citet{Interpretable_CF} investigate forgetting in LLMs through the perspectives of instruction-following and task-related knowledge. They highlight that the forgetting stems from a decline in instruction-following capabilities rather than an actual loss of task-related knowledge. Nevertheless, they employ disparate experimental settings—instruction-following for model training versus prefix completion for knowledge probing—which weakens the persuasion of their conclusions. 

\begin{figure}
    \centering\includegraphics[width=1\linewidth]{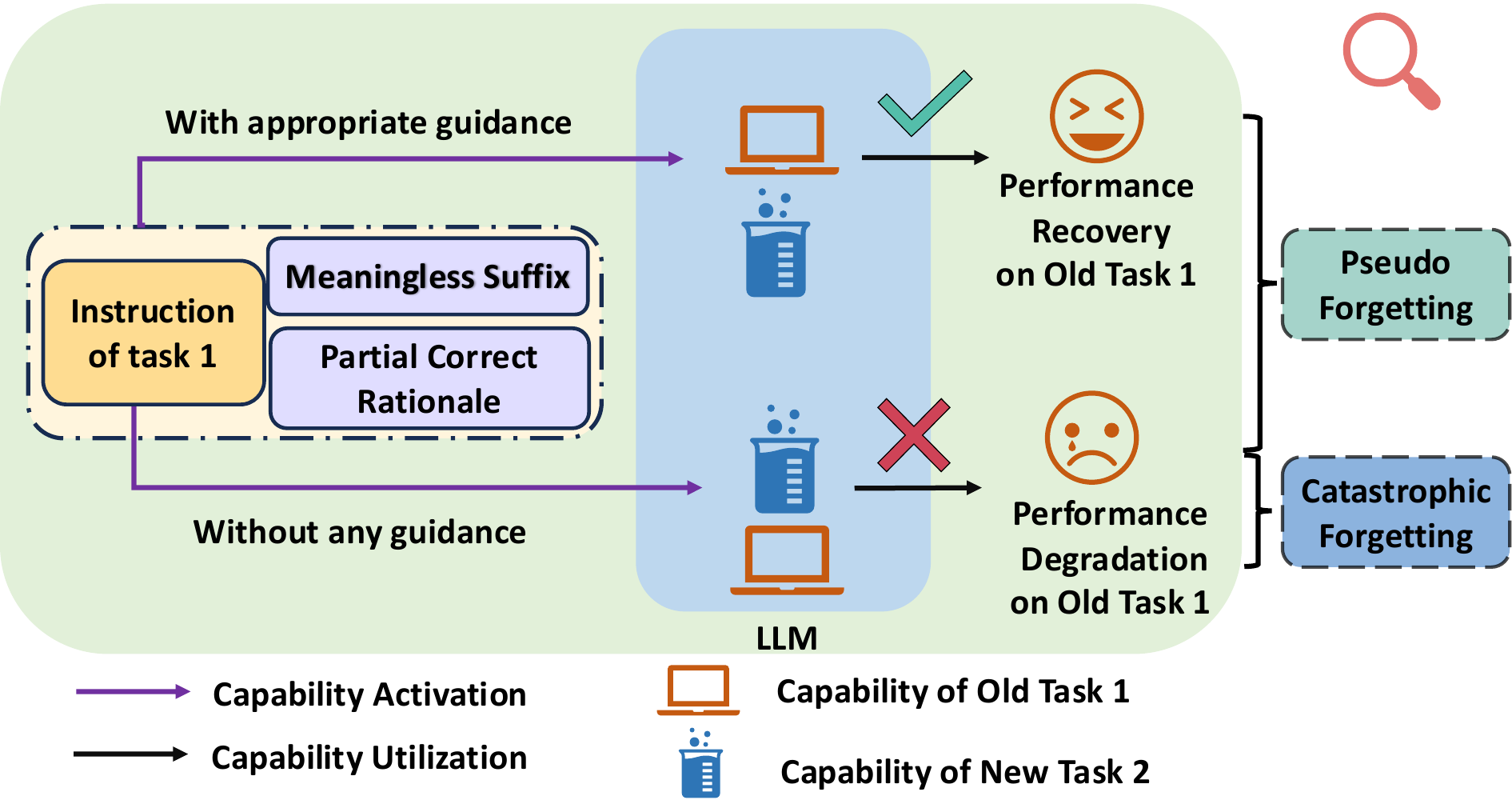}
    \caption{Pseudo forgetting. \textbf{1.} The performance degradation on previous tasks stems from instructions failing to properly activate the model's inherent capabilities rather than genuine forgetting of task-relevant abilities. \textbf{2.} Performance can be restored through appropriate prompting, demonstrating no actual forgetting occurs.}
    \label{fig:pseudo_forgetting}
    \vspace{-10pt}
\end{figure}

In this paper, as shown in Figure~\ref{fig:pseudo_forgetting}, we argue that the observed performance degradation on previous tasks stems not from a genuine loss of task capabilities, but rather from the instructions' failure to effectively activate the model's intrinsic abilities—a phenomenon we term "pseudo forgetting". To validate this hypothesis, we conduct probing experiments on LLMs across a range of natural language tasks under instruction-following settings. We find that, given partial rationale as external guidance or augmented with a task-irrelevant instruction suffix, the forgetting model can complete the rationale and recover performance close to its pre-forgetting level, providing strong empirical support for our hypothesis. To investigate the underlying causes of pseudo forgetting, we employ attribution scores to quantitatively analyze the model’s reliance on the instructions during rationale generation. Our analysis reveals that the pseudo-forgetting model exhibits significantly reduced reliance on instructions, which prevents the model from effectively utilizing its internal capabilities.

Building on the above insights, we believe that when learning new tasks, replaying data related to previous tasks to strengthen the model’s reliance on corresponding instructions offers a simple and effective solution to mitigate pseudo forgetting. However, how to allocate replay data efficiently is limited studied~\citep{InsCL}. Thus, we first introduce the Rationale-Guidance Difficulty (RGD) metric, which measures the model’s ability to leverage the correct internal capability under a given instruction. We then propose Rationale-Guidance Difficulty based Replay (RGD-R) to optimize the data utilization in replay-based continual learning algorithms. Specifically, during continual learning, the RGD score for each previous task is dynamically computed and used to determine the ratio of required replay data. Experimental results demonstrate that RGD-R effectively alleviates pseudo forgetting while preserving the model’s plasticity\footnote{Code and data are available at \href{https://github.com/DIRECT-BIT/Reviving-Dormant-Memories}{here}.}.

Our contributions can be summarized as follows:
\begin{enumerate}
    \item We directly demonstrate the existence of pseudo forgetting in the continual learning of LLMs (Section~\ref{sec:evidence_exp}), followed by an analysis of the underlying cause (Section~ \ref{sec:information_loss_exp}).
    \item Building on this insight, we introduce RGD score, which measures the model’s ability to leverage the correct intrinsic capabilities under a given instruction (Section~\ref{sec:RGD}).
    \item By adopting RGD, we develop RGD-R, a novel replay-based framework designed to maximize the efficiency of replay data via dynamic data allocation (Section~\ref{sec:RDG-R}).
\end{enumerate}

\section{Unveiling Pseudo Forgetting : the evidence and cause}
\label{sec:exp_to_demo_hy}


\begin{tcolorbox}[colframe=gray!80!black, colback=white, coltitle=black, colbacktitle=white, title={\bfseries Pseudo Forgetting}]
Pseudo forgetting is a phenomenon where performance degradation on previously learned tasks in continual learning occurs not through the loss of task capabilities, but rather through the diminished effectiveness of original task instructions in activating the model's intact intrinsic capabilities, resulting in incorrect rationales and outputs.
\end{tcolorbox}

In Section~\ref{sec:evidence_exp}, we directly demonstrate that models do not genuinely forget task capabilities by restoring their performance on previous tasks via employing two methods to provide appropriate guidance. In Section~\ref{sec:information_loss_exp}, we quantify the model's reliance on instructions during rationale generation, revealing that pseudo forgetting occurs because original instructions fail to activate the model's appropriate intrinsic capabilities.
\begin{figure*}[tbp]
    \centering
    \includegraphics[width=1\linewidth]{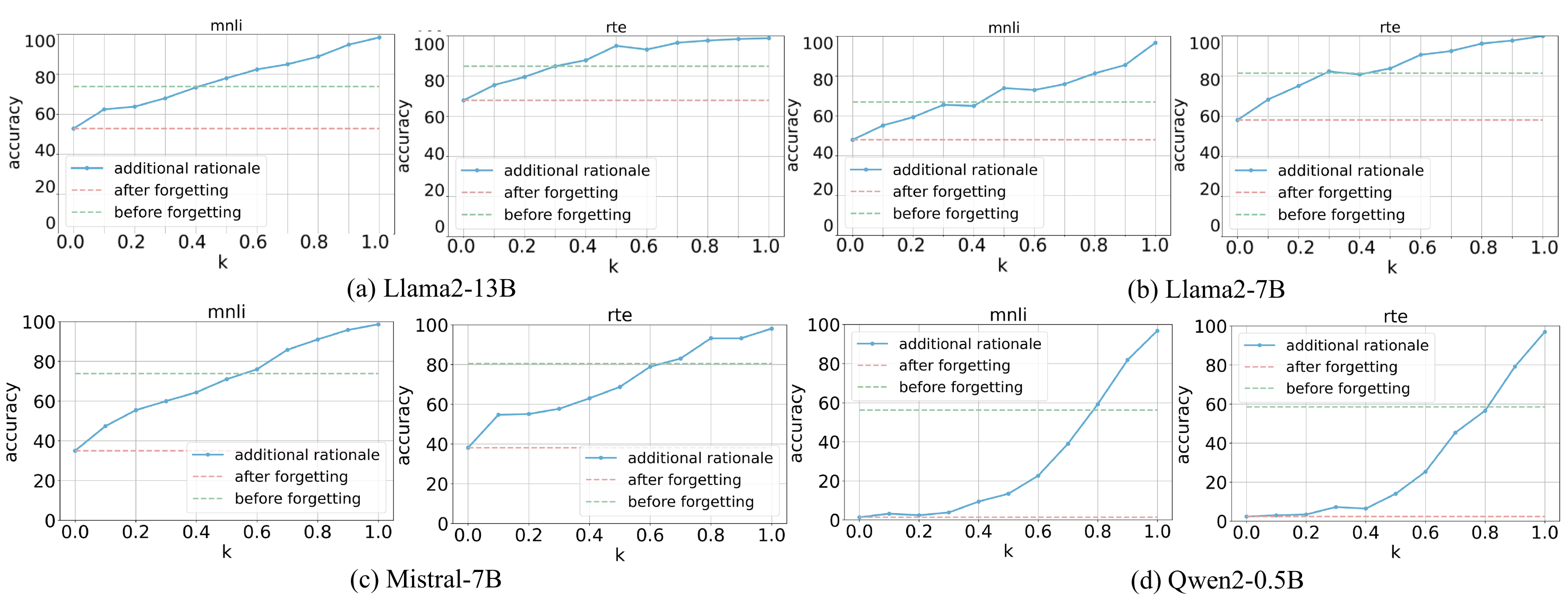}
    \caption{Changes in the model's task performance after forgetting when the first $k$ portions of the appropriate rationale are provided. \textbf{1.} A forgetting model can regenerate the ``forgotten rationale'' and gradually recover its ``pre-forgetting'' task performance when passively guided with partial “appropriate rationale.” \textbf{2.} The degree of recovery of the task performance is related to the task difficulty and the scale of the model.}
\label{fig:ex_knowledge_exp_res_llama}
\end{figure*}
\subsection{Evidence for Pseudo Forgetting}
\label{sec:evidence_exp}
For a forgetting model, two fundamental questions naturally arise:
\begin{enumerate}
    \item \textbf{Q1}: \textit{How does the model perform when \textbf{passively} provided with external correct rationale?} 
    \item \textbf{Q2}: \textit{Can changing prompt (eg. adding task-irrelevant prefixes or suffixes) enable the model to generate the correct rationale \textbf{actively}}?
\end{enumerate}

\subsection*{A1: With a partially correct rationale guidance, the model demonstrates potential for passively recovering task performance.}
\label{exp:A1}
\begin{figure}[h]
    \centering
    \includegraphics[width=1\linewidth]{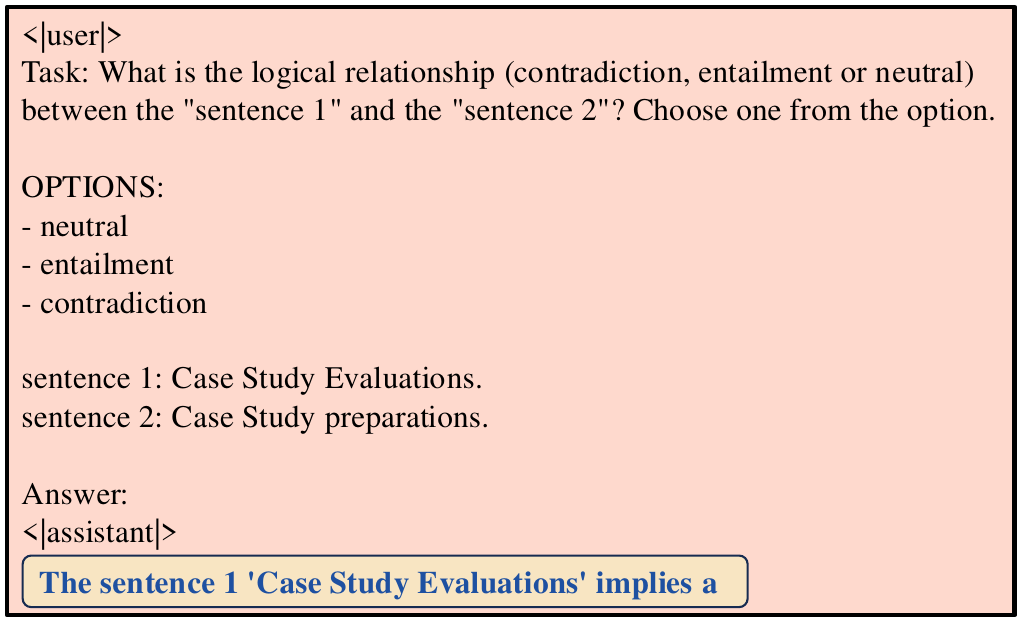}
    \caption{Prompt example with additional the first 10\% words of the correct rationale guidance ($k=0.1$). The black parts are the original instruction;  The blue parts are the added part of the correct rationale, which does not contain information directly related to the answer.}
    \label{fig:ex_knowledge_prompting}
\end{figure}

\paragraph{Experiment Setting} To address \textbf{Q1}, we select the model from the final stage of sequential learning and choose the test set of tasks with a high forgetting rate for this experiment. To offer external correct capability guidance, as shown in Figure~\ref{fig:ex_knowledge_prompting}, the first $k$ portions of the ground truth rationale after the \texttt{<|assistant|>} token, where $k$ is the ratio range from 0 to 1 ($k\in[0,1]$)\footnote{See Appendix~\ref{app:imp_k_portions} for detailed implementation code.}. Notably, our experiments in the Appendix~\ref{app:eval_k_rationale} show that providing a small ratio ($k\le0.2$) of the correct rationale does not directly convey task-critical information to the model, but rather guides the model in shaping the overall direction of its predictions.

\paragraph{Results and Analysis} The result is illustrated in Figure~\ref{fig:ex_knowledge_exp_res_llama}.
Firstly, \textbf{under the guidance of correct rationale, a forgetting model demonstrates promising potential to recover task performance to pre-forgetting levels}. Specifically, the performance on different forgotten tasks improves consistently across varying model scales as the value of $k$ increases. Secondly, \textbf{the potential for recovery of task performance is related to the task difficulty and the scale of the model}. For instance, in the RTE task, Llama2-13B returns to its pre-forgetting performance level at $k=0.3$, while the MNLI task requires $k=0.4$ to achieve the same recovery level. Meanwhile, to restore performance on MNLI and RTE to pre-forgetting levels steadily, Qwen2-0.5B, Mistral-7B, Llama2-7B, and Llama2-13B require $k$ values of 0.8, 0.6, 0.5, and 0.4, respectively.

However, as shown in Table~\ref{tab:evaluation_of_k_rationale} of Appendix~\ref{app:eval_k_rationale}, since the externally provided partial rationales introduce no task-relevant information only when $k\le0.2$, we propose the following two potential explanations for the observed results:
\begin{enumerate}
    \item [(1).] \textbf{Complete catastrophic forgetting}: LLMs require external reasoning guidance to restore performance (even Llama2-13B at $k=0.4$), suggesting they may simply leverage provided solution components rather than retain problem-solving abilities.
    \item [(2).] \textbf{Capability activation failure}: LLMs' improved performance under minimal guidance indicates preserved capabilities, as critical reasoning steps were self-generated rather than externally provided (\textit{e.g. $k=0.1$ in Figure~\ref{fig:ex_knowledge_prompting}}). Specifically, when $k=0.2$, both 13B and 7B scale models demonstrate partial recovery of performance on the forgotten tasks.
\end{enumerate}
To determine which of these two explanations is correct, we conduct further investigation into \textbf{Q2}.
\subsubsection*{A2: With the addition of task-agnostic instruction suffixes, the model can actively recover its original task performance.}
\label{exp:gcg}
\begin{figure}[!h]
    \centering
    \includegraphics[width=1\linewidth]{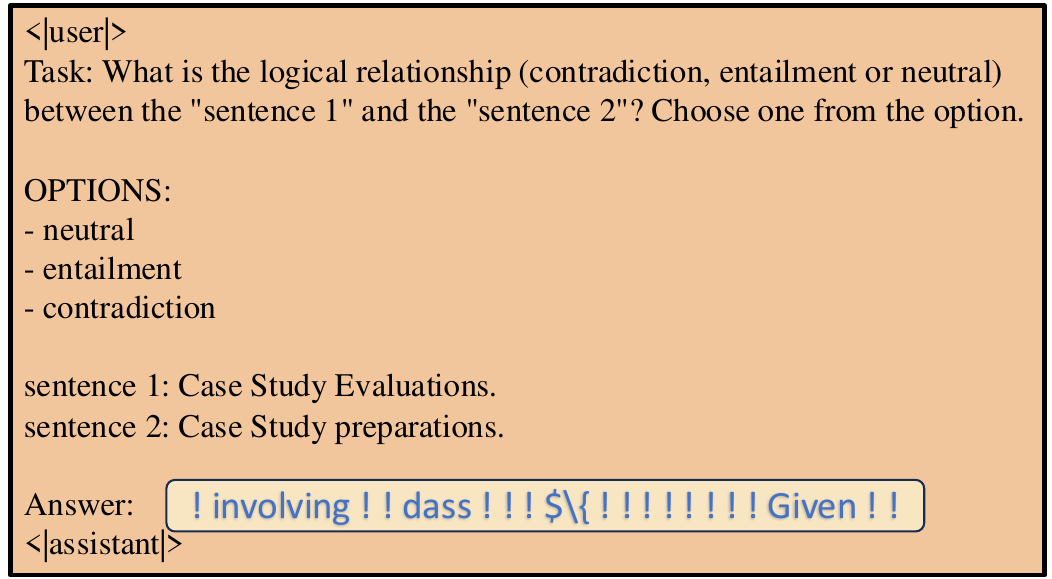}
    \caption{Prompt example with \textcolor{blue}{\textit{task-irrelevant suffix}} searched by Greedy Coordinate Gradient~\citep{zou2023universal}. The forgetting model outputs \textit{Health and Wellness} due to the influence of the previous task, Yahoo, but correctly outputs \textit{entailment} before forgetting or augmenting with this suffix.}
    \label{fig:gcg_aufix}
\end{figure}

 We employ Greedy Coordinate Gradient-based Search~\citep{zou2023universal} to search for a meaningless suffix that helps the original instruction guide the forgetting model toward proper rationale generation actively, as shown in Figure~\ref{fig:gcg_aufix}.

\paragraph{Greedy Coordinate Gradient-based Search (GCG)} 
Given a sequence $x_{1:n}$, the probability of generating a sequence $x_{n+1:n+T}$ can be written as: 
\begin{equation}\small
    p(x_{n+1:n+T} \mid x_{1:n})=\prod^T_{i=1} p(x_{n+i}\mid x_{1:n+i-1})
\end{equation}
Under the above notation, the loss of generating a target sequence $T=x_{1:N_{target}}$ (eg. partial correct rationale) given an instruction $I=x_{1:N_{ins}}$ and an initial suffix $S=x_{1:N_{suffix}}$ can be written as 
\begin{equation}\small
    \mathcal{L}(S)=-\log p(T\mid [I,S])
\end{equation}

To minimize the above loss, GCG~\citep{zou2023universal} leverages gradients with respect to the one-hot token indicators to identify promising token replacements. Specifically, for each token position $i$, in the suffix, the gradient $\nabla \mathcal{L}_{e_i}(S)$ is computed, where $e_i$ is the one-hot vector representing the current token at position $i$. Then, for each token position, the top-$k$ tokens with the largest negative gradients are identified as candidate replacements. Finally, the candidate replacement that minimizes the loss is selected and applied to the suffix.


Notably, this approach ensures the validity of the experiments: (1) semantically meaningless suffixes devoid of task-specific information, ensuring the generated rationale reflects parametric capabilities; (2) instruction-following setting remains unchanged, aligning the detected capabilities with those learned via instruction fine-tuning, in contrast to the probing experiments in~\citet{Interpretable_CF}, which is under prefix completion setting.



\paragraph{Experimental Settings}
We evaluate models from the final stage of sequential learning ($M_{a-f}$). For each task, we sample 100 instances where models exhibit correct predictions before forgetting but fail after forgetting, represented as $D_{\texttt{f}}=\{(I_i,A_i)\}$. For GCG, as shown in Table~\ref{tab:GCG_targets}, we explore three optimization targets $T$: (1) Answer guidance; (2) Partial ground truth rationale guidance; (3) Partial pre-forgetting rationale guidance, where the rationale is generated by the model before forgetting. 
To prevent the incorporation of task-specific information provided by (2) and (3), we restrict the search target to only the first 20\% of the rationale, \textit{i.e.}, $k=0.2$.
The suffix searched for each sample $(I_i,A_i)$ is denoted as $S_i$. See Appendix~\ref{sec:gcg_imp} for the detailed implementation. 

\paragraph{Evaluation Metric} To quantitatively evaluate the extent of task performance recovery of the forgetting model on the forgotten task, we formally define the \textbf{recovery rate (R.Ra)} computed as follows:
\begin{align}\small
\texttt{R.Ra}=\frac1{|D_{\texttt{f}}|}\sum_{I_i,A_i\in D_{\texttt{f}}}\mathbb{I}(M_{a-f}([I_i,S_i]),A_i)
\end{align}
where $\mathbb{I}(M_{a-f}([I_i,S_i]),A_i)$ is an indicator function that equals 1 when $M_{a-f}$ predicts correctly and 0 when it predicts incorrectly.

\paragraph{Results and Analysis}

\begin{figure}[t]
    \centering
    \includegraphics[width=\linewidth]{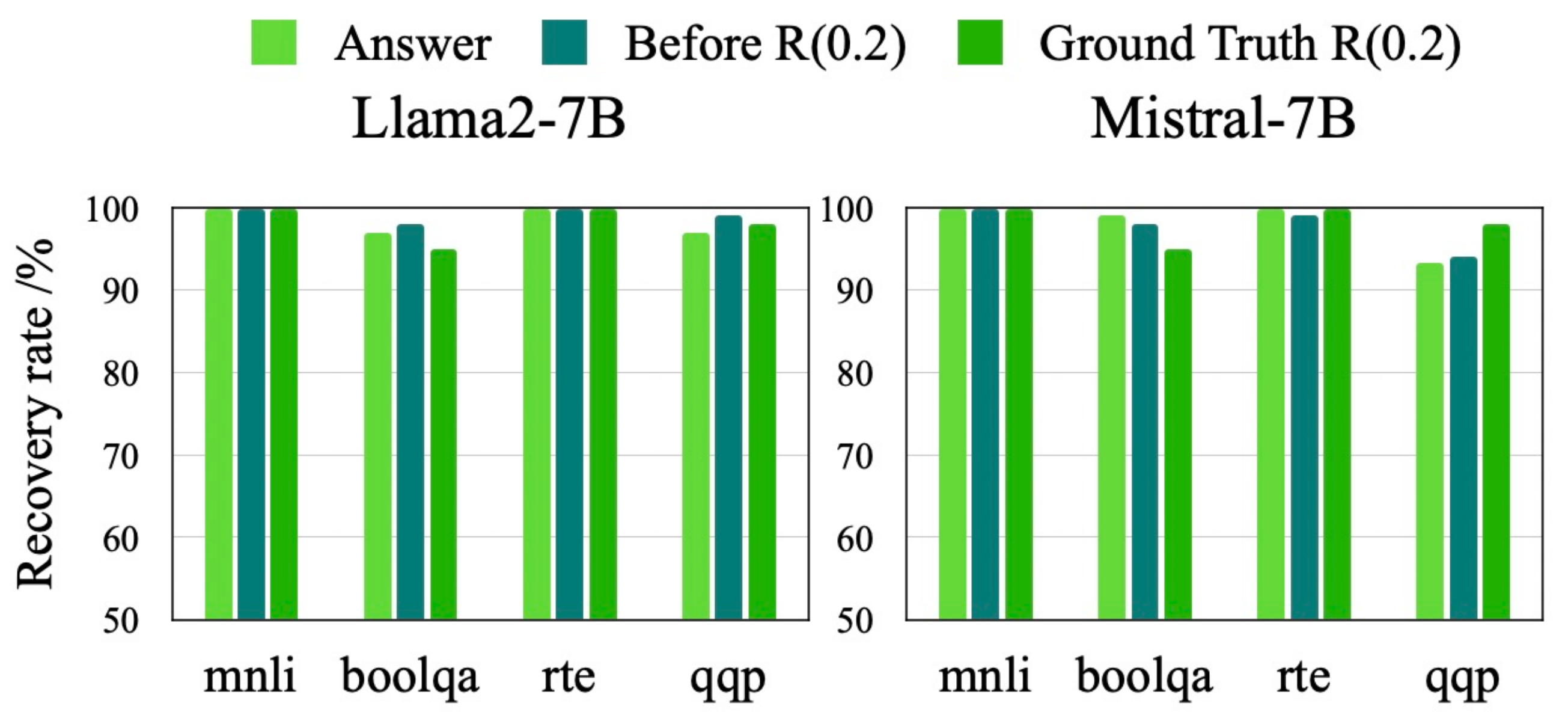}
    \caption{Recovery rate of forgotten tasks. \textbf{1.} For each task, we sample 100 forgotten instances. \textbf{2.} The labels `Answer', `Before R (0.2)', and `Ground Truth R (0.2)' denote respectively: the ground truth answer,  the first 0.2 portions of the rationale generated by the model before forgetting, and the ground truth, serving as optimization target for GCG. \textbf{3.} The recovery rates of different models on various tasks surpass 90\% (even reaching 100\% in specific tasks), indicating the forgetting models preserve previously acquired capabilities.}
    \label{fig:task_agnost_prompting_res}
\end{figure}

As shown in Figure~\ref{fig:task_agnost_prompting_res}, appending task-irrelevant suffixes to original instructions enables forgetting models to actively generate correct rationale, leading to 90\% recovery rate across tasks. This provides direct evidence that the model dose not forget previously acquired capabilities. Specifically, the recovery effectiveness may correlate with sample complexity. While Mistral-7B demonstrates complete recovery (100\%) on MNLI, its average recovery rate on QQP is 95.44\%, with a similar trend observed in Llama2-7B. Table~\ref{tab:Explored_suffix_cases} presents suffix cases searched via GCG based on different models and test samples.

\begin{tcolorbox}[colframe=gray!80!black, colback=white, coltitle=black, colbacktitle=white, title={\bfseries Summary}]
The results of the two experiments provide direct evidence of pseudo forgetting: the model does not truly forget task-specific capabilities, rather, the original instructions fail to guide the model in leveraging the appropriate abilities to solve the task.
\end{tcolorbox}

\subsection{Cause of Pseudo Forgetting}
\label{sec:information_loss_exp}

In this section, we investigate the cause of pseudo forgetting to further validate our hypothesis. We demonstrate that the pseudo-forgetting model exhibits a reduced reliance on the original instructions during rationale generation, preventing the model from correctly leveraging its intrinsic capabilities.

\paragraph{Attribution Algorithm}
We use attribution scores~\citep{rationale_drift, wang-etal-2023-label,dai-etal-2022-knowledge} to quantify and analyze the extent to which the model relies on instructions during the rationale generation stage. Formally, let $Q^{(l)}_{IR}$ denote the dependency score at layer $l$, capturing the dependency
between the instruction $I$ and the rationale $R$. The detailed algorithmic description and implementation are provided in Appendix~\ref{sec:attribution_implementation}. 

\begin{figure}[t]
	\centering
	\subfigure[Llama2-7B] {\includegraphics[width=0.22\textwidth]{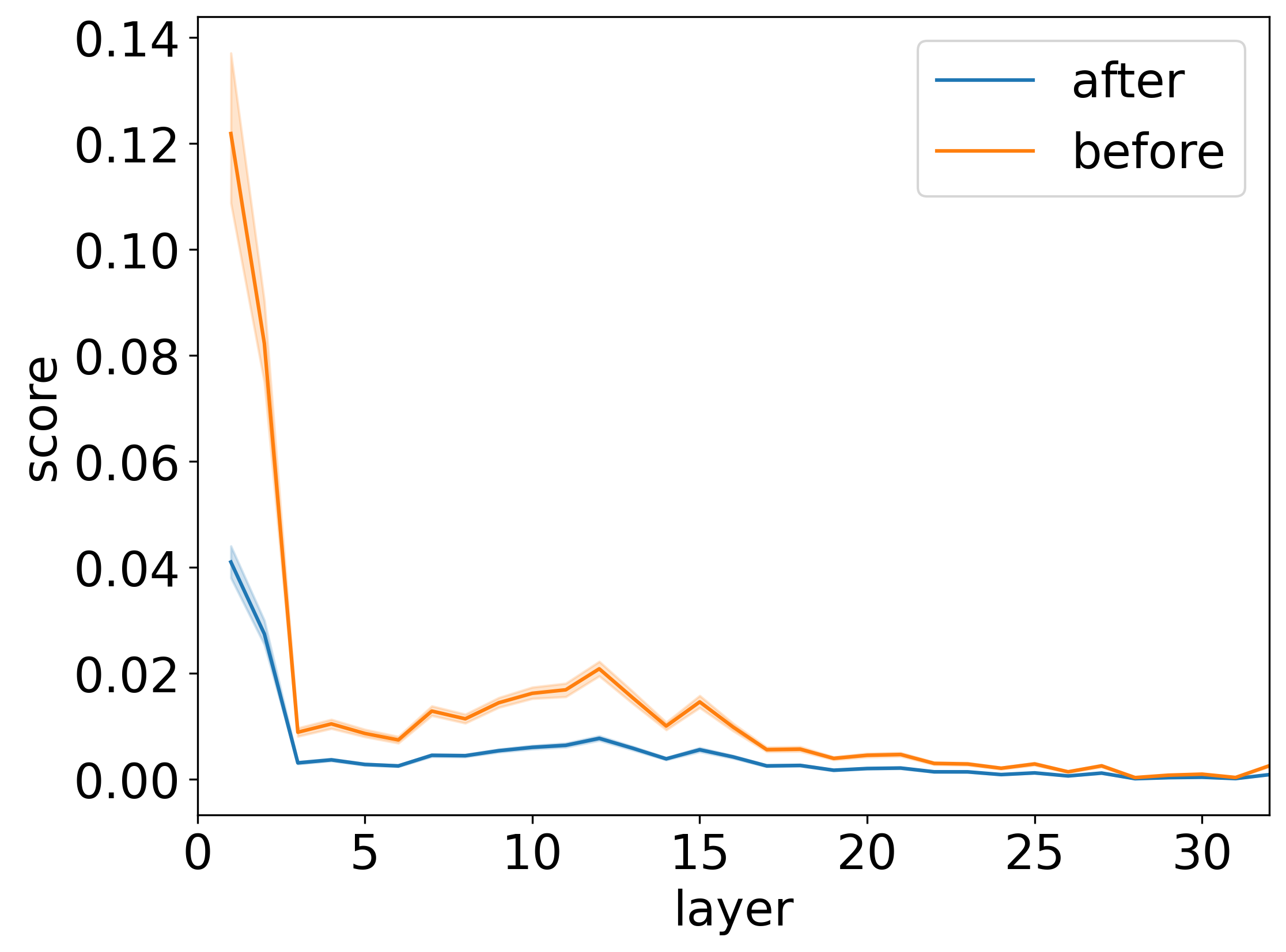}}
	\subfigure[Mistral-7B] {\includegraphics[width=0.23\textwidth]{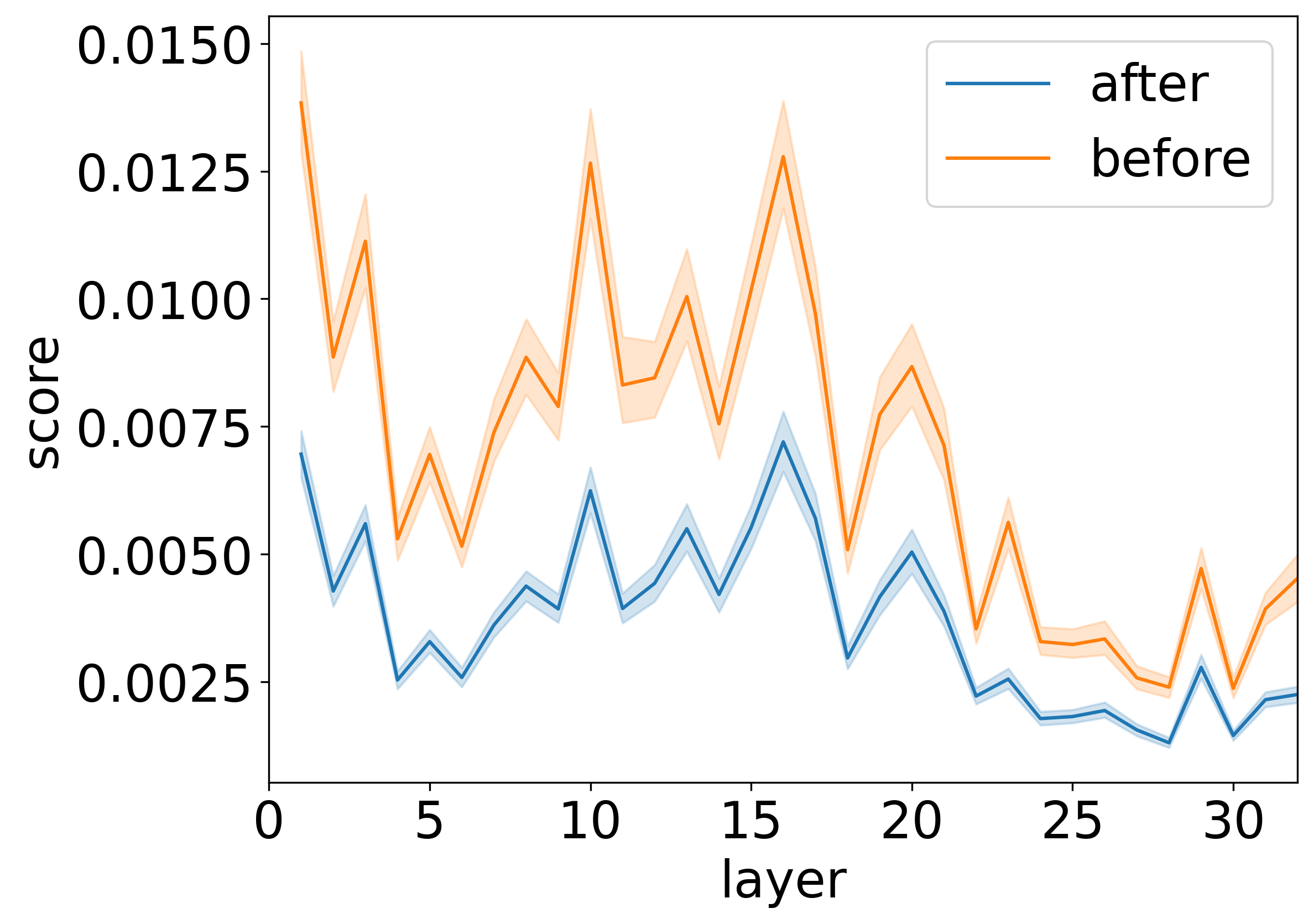}}
	\caption{Comparison of instruction dependency scores of pseudo-forgetting model for generating correct and incorrect rationales on MNLI task.} 
	\label{fig:pre_scre_res}
\end{figure}

\paragraph{Experimental Settings}

We use $M_{b-f}$ and $M_{a-f}$ to denote the model trained on the old task and continually trained on the final task, corresponding to the stages of before and after pseudo forgetting. The probing dataset is the same as that used in Section~\ref{sec:evidence_exp}. Each sample can be denoted as $(I, R_{b-f}, R_{a-f}, R_{g}, A_{b-f}, A_{a-f}, A_{g})$, where $I$ represents the instruction, $R_{b-f}, R_{a-f}, R_{g}$ represent the rationale generated by $M_{b-f}$, $M_{a-f}$, and Llama3.1-70B-Instruct (as the ground truth), respectively. $A_{b-f}, A_{a-f}, A_{g}$ represent the corresponding predicted answers. 

\paragraph{Experiment 1} Firstly, we investigate the differences in the pseudo-forgetting model’s ($M_{a-f}$) instruction dependency when generating incorrect ($R_{a-f}$) versus correct ($R_{a-f}$) rationale.

As shown in Figure~\ref{fig:pre_scre_res} and Figure~\ref{fig:more_pred_attri_on_rte}, we can conclude that \textbf{the pseudo-forgetting model generates incorrect rationales primarily due to the reduced instruction dependency}. Specifically, for $M_{a-f}$, the instruction dependency when generating incorrect rationales (blue line) is generally lower than that when generating correct rationales (orange line). The difference is noticeable in shallow layers, aligning with the findings in \citet{wu-etal-2024-language} that shallow layers learn more and stronger instruction-following patterns.

\paragraph{Experiment 2} Secondly, to confirm that the reduced instruction dependency is indeed caused by pseudo forgetting, we examine the impact of different models ($M_{b-f}$ vs $M_{a-f}$). Specifically, we compare the relative instruction dependency scores when different models generate rationales:
\begin{equation}
\label{eq_delta}
    \Delta_{\text{Attr}(R_{gen}|R_{g})} = |Q^{(l)}_{IR_{gen}} - Q^{(l)}_{IR_{g}}|
\end{equation}
where $R_{gen}$ is $R_{a-f}$ ($R_{b-f}$) if we calculate Equation (\ref{eq_delta}) on $M_{a-f}$ ($R_{b-f}$). This approach ensures that the only variable in the experiment is the occurrence of pseudo forgetting.

\begin{figure}[t]
	\centering
	\subfigure[Llama2-7B] {\includegraphics[width=0.22\textwidth]{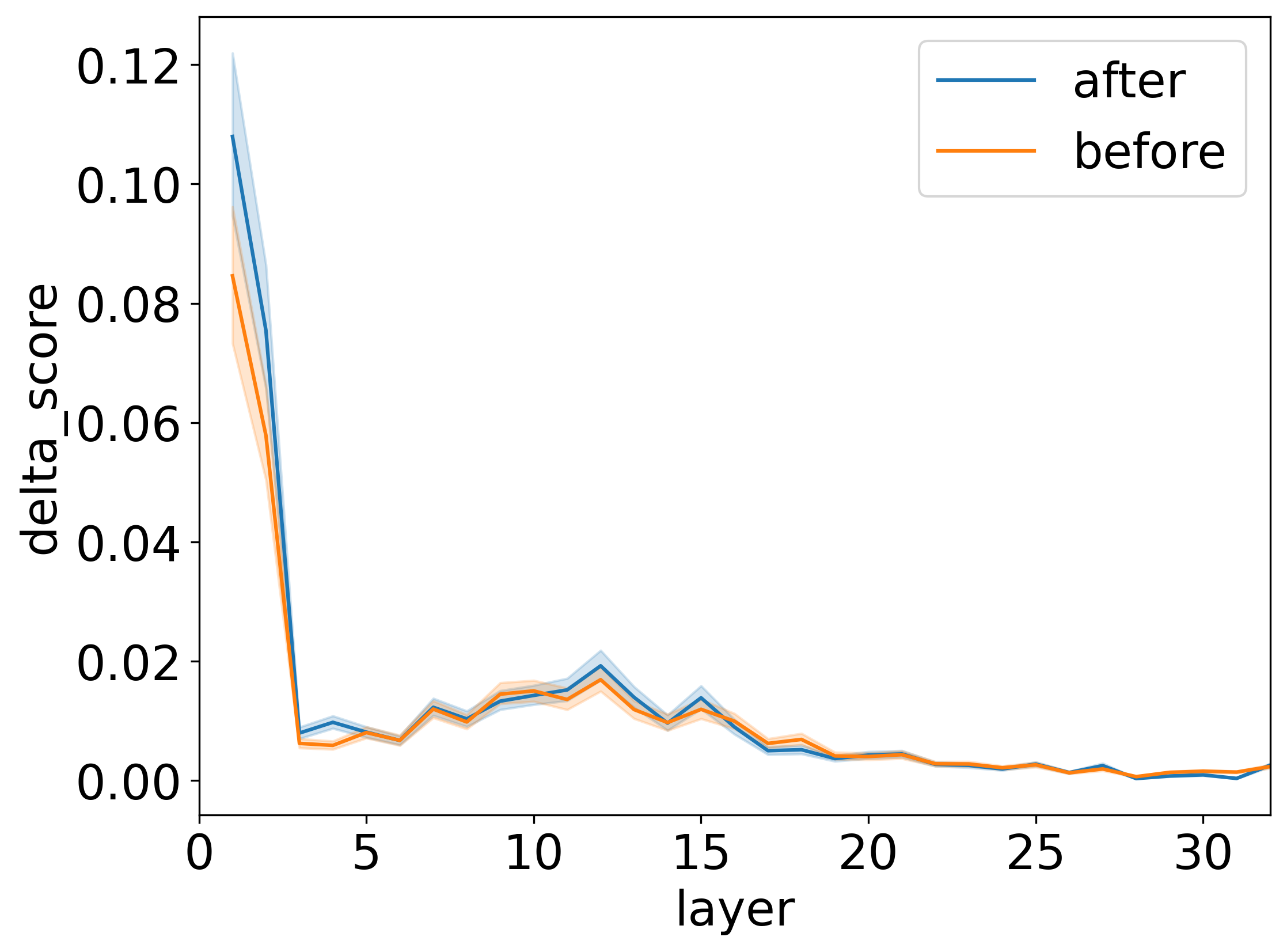}}
	\subfigure[Mistral-7B] {\includegraphics[width=0.23\textwidth]{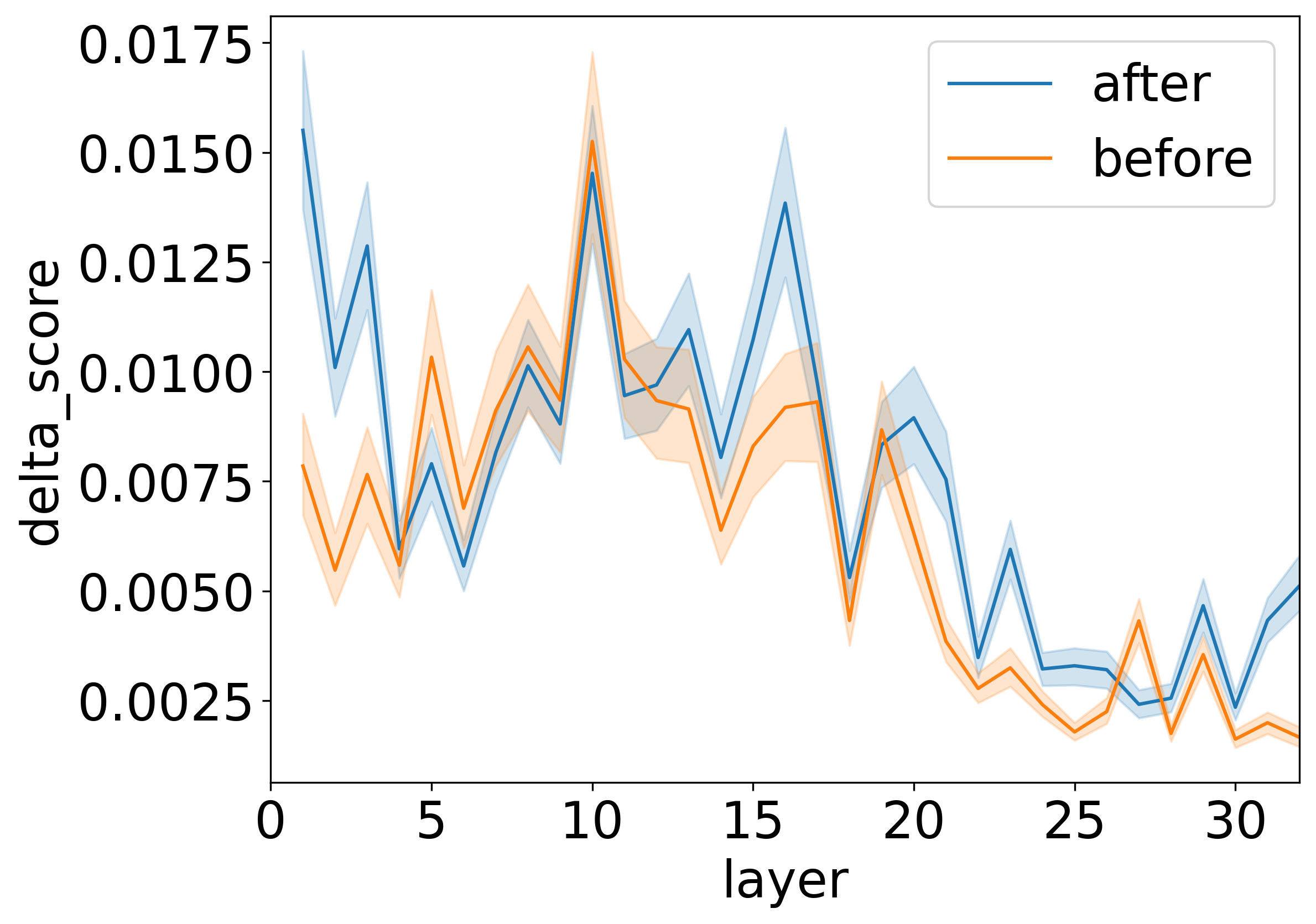}}
	\caption{Comparison of relative instruction dependency scores across different states of Llama2-7B and Mistral-7B on MNLI task.} 
	\label{fig:delta_scre_res}
\end{figure}

As shown in Figure~\ref{fig:delta_scre_res} and Figure~\ref{fig:more_pred_delta_attri_on_rte}, the discrepancy between $R_{g}$ and $R_{a-f}$ on $M_{a-f}$ (blue line) is larger compared to the difference between $R_{g}$ and $R_{b-f}$ on $M_{b-f}$ (orange line). This finding further supports our hypothesis that \textbf{a key factor contributing to pseudo forgetting is the model’s reduced reliance on the original instruction during rationale generation}. While certain layers display differences or larger “before” delta scores compared to the “after” condition, we leave the analysis of this observation to future work.

\section{Addressing Pseudo Forgetting: Rationale-Guidance Difficulty based Replay}
\label{sec:our_method}

Based on these findings, we argue that replay-based algorithms, which incorporate a small portion of data from previous tasks during continual learning, can effectively reinforce the model's dependency on corresponding instructions, thereby offering a simple yet effective solution to pseudo forgetting. However, how to allocate the replay data ratio for each task remains underexplored~\citep{InsCL}. Thus, in Section~\ref{sec:RGD}, we introduce the Rationale-Guidance Difficulty (RGD) metric to measure the impact of pseudo forgetting on the model. Then, in Section~\ref{sec:RDG-R}, we propose Rationale-Guidance Difficulty based Replay (RGD-R), which leverages RGD to dynamically determine the replay data proportion for each task, optimizing replay data utilization during continual learning.

\subsection{Rationale-Guidance Difficulty}
\label{sec:RGD}

We first introduce the Rationale-Guidance Difficulty (RGD) metric, which measures the difficulty for the model to correctly utilize its internal capabilities in generating appropriate rationale under a given instruction. A higher RGD score signifies greater difficulty for a prompt in guiding the model to generate the correct rationale, and vice versa. For a data triplet $(I,R_g,A_g)$, the RGD score is calculated as follows:
\begin{equation}\small
    \texttt{RGD}(I, R_{g}, A_{g})=\frac{\texttt{PPL}_{a-f}(R_{g}|I)}{\texttt{PPL}_{b-f}(R_{g})},
\end{equation}
where $I$, $R_{g}$, and $A_{g}$ denote the prompt, the ground truth rationale, and the ground truth answer, respectively. 
$\texttt{PPL}_{b-f}(R_{g})$ represents the difficulty for the model with normal access to its capabilities to generate the correct rationale, and $\texttt{PPL}_{a-f}(R_{g}|I)$ denotes the difficulty for the pseudo-forgetting model to generate the same rationale given prompt $I$. 

\begin{equation}\small
\label{eq:task_rgd}
    \texttt{RGD}_{D} = \frac{1}{|D|}\sum_i \texttt{RGD}
    {(I,R_g,A_g)_i},
\end{equation}
where $(I,R_g,A_g)_i$ is the $i$-th sample in dataset $D$, and $|D|$ is the total number of samples.

\subsection{Theoretical Analysis}


Here, we give a simple proof that under a reasonable assumption, the RGD score can measure the difficulty of the capability activation process. First, \citet{wu-etal-2024-language} finds that the underlying mechanism of instruction following likely involves model $\theta$ first recognizing instruction $i$, then utilizing the activated capabilities $c_1,\dots,c_n$ to generate rationale $r$. We can formalize this process as:
\begin{equation}\small
\label{eq_p_i_r}
     P_\theta(r|i) = \sum_n p(r\mid c_n)\cdot p(c_n\mid i).
\end{equation}
%

\begin{assumption} (Independence of Task Abilities)
\label{asp:2}
Under normal circumstances, each capability $c$ can only be activated by task-specific instruction $i$, which subsequently supports the generation of the corresponding rationale $r$. The capabilities of tasks across different domains are independent from one another. This can be formulated as:
\begin{align}
\label{eq:assump}
   \forall m \neq n, \quad p(r \mid c_n) \cdot p(c_m \mid i) = 0.
\end{align}
\end{assumption}

Given this assumption, we can formalize the probability of the model $\theta$ to activate the correct task capability $c^*$ given instruction $i$ as:
\begin{equation}\small
    P_\theta(c^* \mid i)=p(c_1,\dots,c_m \mid i)= \sum_m p(c_m \mid i),
\end{equation}
 and the process of generating the correct rationale $r^*$ based on the model’s internally activated capabilities can be formally expressed as follows:
:
\begin{equation}\small
    P_\theta(r^*)=p(r^*\mid c_1,\dots,c_n)= \sum_n p(r^* \mid c_n)
\end{equation}
Given Equation (\ref{eq:assump}), we can rewrite Equation (\ref{eq_p_i_r}) as:
\begin{equation}\small
    P_\theta(r^*|i) = \big(\sum_n p(r^* \mid c_n)\big) \cdot \big(\sum_m p(c_m \mid i)\big)
\end{equation}
Hence, the following equation holds:
\begin{equation}\small
    P_\theta(c^*\mid i)=\frac{P_\theta(r^*|i)}{P_\theta(r^*)}
\end{equation}

Consequently, following the same principle, the RGD score can approximate the difficulty of a given instruction in activating the model's correct capability to generate the corresponding rationale. 
\subsection{RGD-based Replay framework}
\label{sec:RDG-R}

To optimize the data utilization in replay-based methods, we propose the Rationale-Guidance Difficulty-based Replay (RGD-R) framework. During continual learning, RGD-R dynamically determines the required replay data ratio for each previous task based on the RGD score calculated via Equation (\ref{eq:task_rgd}). Specifically, when training the model on the $i$-th task, the replay data ratio for the $j$-th previous task can be calculated as:
\begin{equation}\small
    \alpha_j= \frac{\texttt{RGD}_{D_j}}{\sum_{k=1}^{i-1}\texttt{RGD}_{D_k}}, \quad j \in [1, i-1]
\end{equation}
where $\sum_{j=1}^{i-1}\alpha_j=1$, and $\texttt{RGD}_{D_j}$ represents the RGD score of the $j$-th previous task. Thus, the amount of replay data allocated to this task is $\alpha_j\cdot N$, where $N$ represents the total amount of replay data.

In the RGD-R framework, tasks that suffer more severely from pseudo forgetting are replayed with more training samples. This adaptive strategy facilitates the recovery of the model's dependency on the corresponding instructions, enabling more effective utilization of the correct task-specific capabilities.

\subsection{Experiments}
\label{sec:all_exps}
\label{Exp_Setting}
\subsubsection{Experiment Setting}
\paragraph{Datasets}  
Following~\citet{razdaibiedina2023progressive} and \citet{wang-etal-2023-orthogonal}, we conduct experiments on Long Sequence Benchmark, with train/validation/test splits of 1000/500/500 samples respectively. See Appendix~\ref{sec:dataset_details} for more details.

\paragraph{Metrics} Following prior works~\citep{SAPT,zhang-etal-2023-citb}
Let $a_{i,j}$ be the testing performance on the $j$-th task after training on $i$-th task, the metrics for evaluating are: (1) \textbf{Final Average Performance (FAP)} is the average performance of all tasks after the final task $t_T$ is learned, i.e., $\texttt{FAP}_T=\frac{1}{T}\sum_{t=1}^T a_{T,t}$; (2) \textbf{Forgetting Rate (F.Ra)} measures how much knowledge has been forgotten across the first $T-1$ tasks, i.e., $\text{F}_T=\frac{1}{T-1}\sum^{T-1}_{t=1}(\max_{k=i}^{T-1}a_{k,t}-a_{T,t})$; (3) \textbf{Backward Transfer (BWT)} measures the impact that continually learning on subsequent tasks has on previous tasks, i.e., $\texttt{BWT}_T=\frac 1 {T-1}\sum^{T-1}_{t=1}(a_{T,t}-a_{t,t})$. (4) \textbf{Forward Transfer (FWT)} measures how much the model can help to generalize and learn the new task, i.e., $FWT = \frac 1T\sum ^T_{t=2}a_{t-1,t}$. Better scores on FAP, F.Ra, and BWT indicate improved model stability, while a better FWT score reflects enhanced model plasticity.

\paragraph{Baselines}
To validate the effectiveness of RGD in measuring pseudo forgetting and RGD-R in mitigating this phenomenon, we conduct comparative experiments across the following baselines focusing on replay data allocation, where samples for each task are randomly selected from the training set: (1) \textbf{Sequential Training (SEQ)} refers to learning new capabilities without replay data; (2) \textbf{Equal Allocation (EA)} replays the same amount of data for each previous task. More training details are provided in Appendix~\ref{sec:Training_Details}.

\begin{table}[t]\small
    \centering
    \begin{tabular}{lrrrr}
    \toprule
    \bf Method &\bf FAP$\uparrow$&\bf F.Ra$\downarrow$&\bf BWT$\uparrow$ &\bf FWT$\uparrow$\\
    \midrule
    \rowcolor{mygray}\multicolumn{5}{c}{\textbf{\textit{Qwen2-0.5B}}}\\
    SEQ&20.73&53.18&-53.04&21.46\\
    EA&64.13&5.43&-4.90&\textbf{33.34}\\
    RGD-R&\textbf{65.99}&\textbf{3.64}&\textbf{-3.29}&31.87\\
    \midrule

    \rowcolor{mygray}\multicolumn{5}{c}{\textbf{\textit{Qwen2-7B}}}\\
    SEQ&70.97&11.78&-11.68&67.53\\
    EA&78.34&3.84&-2.67&\textbf{69.87}\\
    RGD-R&\textbf{79.76}&\textbf{2.21}&\textbf{-1.05}&69.63\\
    \midrule
    \rowcolor{mygray}\multicolumn{5}{c}{\textbf{\textit{Mistral-7B}}}\\
         SEQ&51.48&30.19&-29.97&47.91\\
         EA&72.15&7.59&-6.96&\textbf{51.17}\\
         RGD-R&\textbf{74.91}&\textbf{4.37}&\textbf{-3.92}&50.77\\
    \midrule
    \rowcolor{mygray}\multicolumn{5}{c}{\textbf{\textit{Llama2-7B}}}\\
    SEQ&62.79&17.87&-17.85&43.95\\
    EA&76.10 &3.52&-2.49&50.91\\
    RGD-R&\textbf{77.03}&\textbf{2.65}&\textbf{-1.25}&\textbf{51.06}\\
    \midrule
    \rowcolor{mygray}\multicolumn{5}{c}{\textbf{\textit{Llama2-13B}}}\\
    SEQ&68.38&13.54&-13.2&51.69\\
    EA&76.98&4.73&-3.70&56.92\\
    RGD-R&\textbf{78.25}&\textbf{3.68}&\textbf{-2.29}&\textbf{57.83}\\
    \bottomrule
    \end{tabular}
    \caption{Performance of different models on Long Sequence Benchmark. The decoding strategy is greedy search. RGD-R effectively alleviates model forgetting and maintains model plasticity simultaneously.}
    \label{tab:all_exp_res_long_seq}
    \vspace{-10pt}
\end{table}

\subsubsection{Main Results}
\textbf{LLMs exhibit inherent resistance to pseudo forgetting, which improves with larger model sizes.} Larger models show lower forgetting rates, such as F.Ra of Qwen2-7B and Qwen2-0.5B with SEQ are 11.78 and 53.18, respectively. 

\textbf{The equal allocation method significantly alleviates pseudo forgetting.} 
Compared to SEQ, EA improves the final performance (FAP) of Qwen2-0.5B, Mistral-7B, and Llama2-13B by 43.40, 20.67, and 8.60, respectively, while reducing the forgetting rate (F.Ra) by 49.54, 22.6, and 9.86. These results support our hypothesis that LLMs do not truly forget the previously learned capabilities.

\textbf{RGD-R further alleviates pseudo forgetting and ensures the model plasticity simultaneously.} 
Compared to EA, RGD-R demonstrates superior effectiveness in mitigating pseudo forgetting (FAP, F.Ra, BWT) and promoting asynchronous knowledge transfer (FWT) across different models. This highlights the efficacy of the RGD score in measuring the impact of pseudo forgetting and confirms that RGD-R successfully optimizes the utilization of replay data in replay-based continual learning algorithms, leading to better overall performance.

\subsubsection{Analysis}
\paragraph{Data Replay Restores Instruction Dependence}

\begin{figure}[t]
	\centering
	\subfigure[MNLI] {\includegraphics[width=0.23\textwidth]{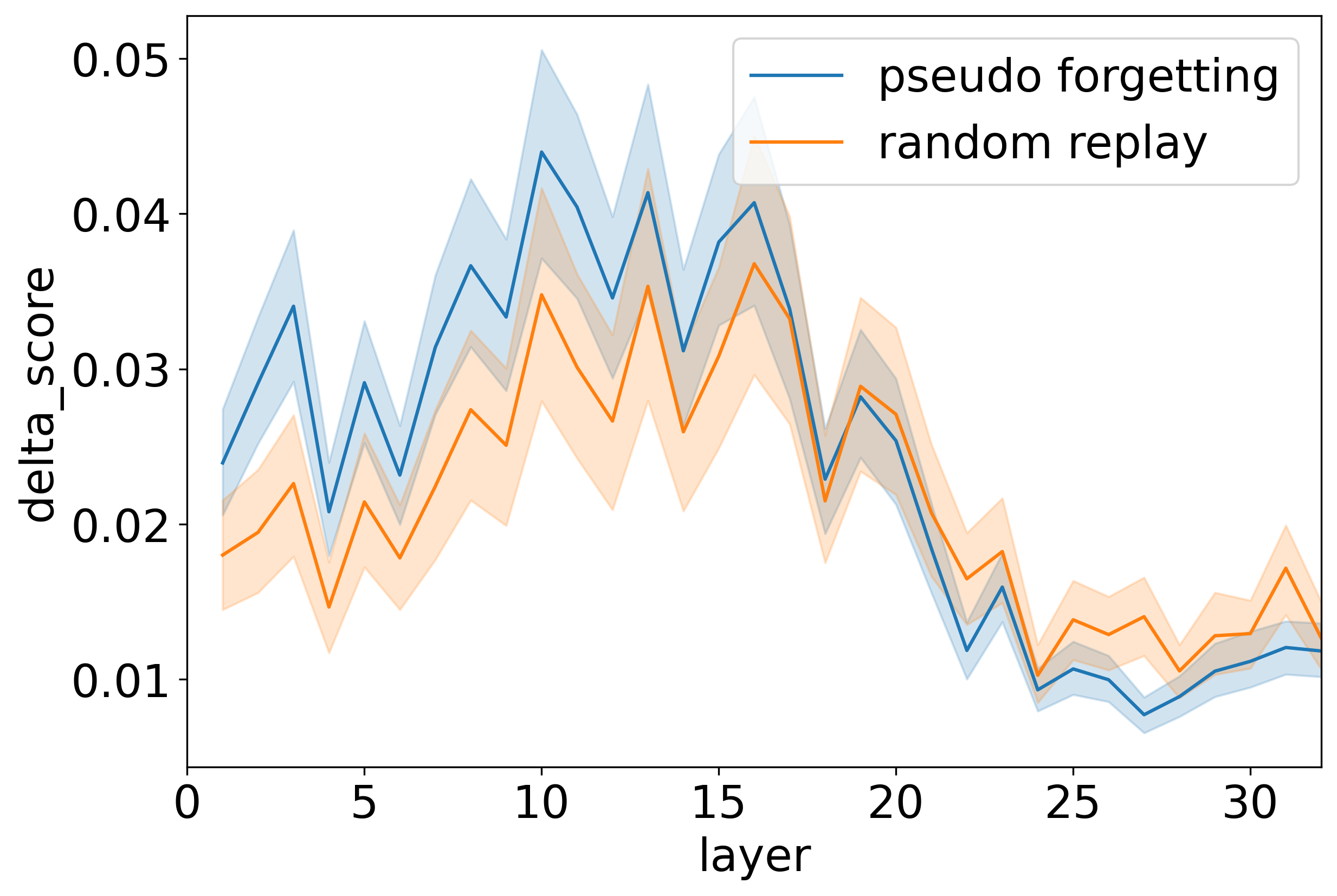}}
	\subfigure[QQP] {\includegraphics[width=0.23\textwidth]{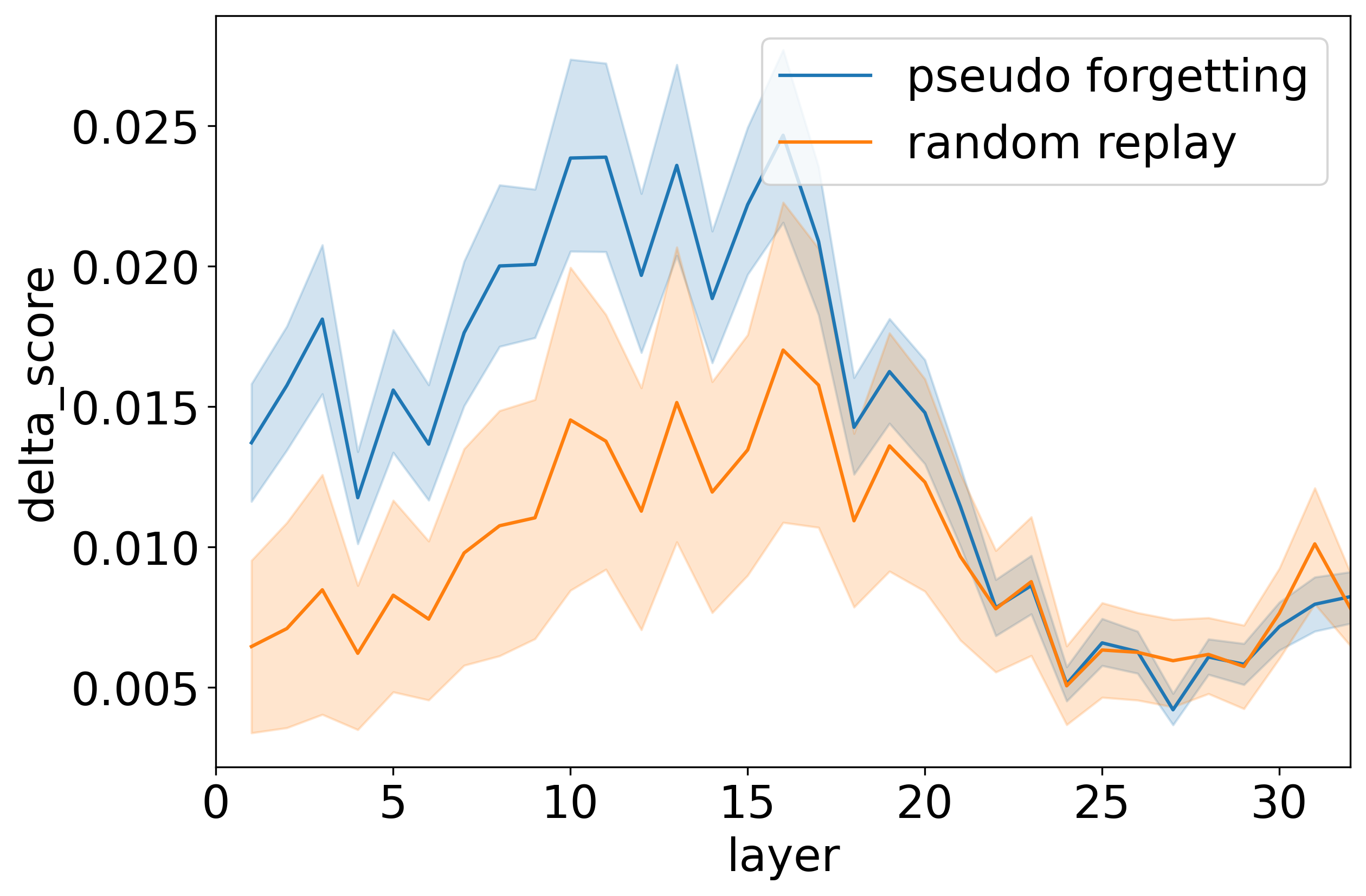}}
	\caption{Comparison of relative instruction dependency scores across different states of Mistral-7B on MNLI and QQP tasks. \textbf{1.} `pseudo forgetting' and `random replay' represent Mistral-7B exhibiting pseudo forgetting and Mistral-7B after capability recovery through random data replay, respectively. \textbf{2.} The replay-based method leads to lower relative instruction dependency scores, indicating that it helps the model rely more on instructions during rationale generation.} \label{fig:_recover_delta_score_res}
\end{figure}

To demonstrate that the replay-based method indeed enhances the instruction dependence, we repeat the attribution experiment in Section~\ref{sec:information_loss_exp}. Specifically, we compare the relative instruction dependency scores between the pseudo-forgetting model trained via SEQ and the model trained via EA data replay. As shown in Figure~\ref{fig:_recover_delta_score_res}, the model trained via data replay (orange line) exhibits a smaller overall difference in instruction dependence when generating rationales compared to the pseudo-forgetting model (blue lines). This suggests that the replay-based method improves the model’s reliance on original instructions, thereby alleviating pseudo forgetting.

\paragraph{Data Replay Enables Better Semantic Recovery in Rationales}
\begin{table}[h]
    \centering
    \begin{tabular}{lccc}
    \toprule
         \textbf{Rationale}&\bf MNLI& \bf BOOLQA &\bf RTE\\
    \midrule
        $R_{a-f}$ &0.2756 &0.2962 &0.2538 \\
        $R_{Paraphrase}$ &0.6641 &0.6793 &0.6554 \\
        \hdashline
        $R_{GCG}$ & 0.2871 &0.3134 & 0.2856\\
        $R_{g}$ &0.4103 &0.4719 &0.4038 \\
        $R_{Replay}$ &\textbf{0.4391} &\textbf{0.4931} &\textbf{0.4359}\\
    \bottomrule
    \end{tabular}
    \caption{Comparison of ROUGE-L scores between rationales ($R_{(\cdot)}$) generated by different methods and those ($R_{b-f}$) from the model before pseudo-forgetting. $R_{Paraphrase}$ is the paraphrased rationale generated by GPT-3.5 based on $R_{b-f}$. $R_{GCG}$ and $R_{Replay}$ are the rationales generated after mitigating pseudo forgetting with the GCG and data replay methods, respectively.}
    \label{tab:Semantic_Compair}
\end{table}


Here, we provide additional evidence from the semantic perspective of rationales, demonstrating that the replay-based method offers a simple yet superior choice. We compare the semantic similarity between rationales generated by different methods ($R_{(\cdot)}$) and those generated by the pre-pseudo-forgetting model ($R_{a-f}$). As shown in Table~\ref{tab:Semantic_Compair}, the replay-based method achieves higher semantic similarity compared to GCG, and surpasses the ground truth rationales. This indicates that replay-based methods are more effective in stimulating the model’s previously learned task capabilities. In contrast, based on GCG, the pseudo-forgetting model still tends to generate tokens related to the new task~\citep{gu-feng-2020-investigating}. While adding a semantic constraint to GCG helps alleviate this issue, our preliminary experiments show that it makes the search process harder and less efficient.

\begin{table}[h]\small
    \centering
    \begin{tabular}{lrrrr}
    \toprule
    \bf Method &\bf FAP$\uparrow$&\bf F.Ra$\downarrow$&\bf BWT$\uparrow$ &\bf FWT$\uparrow$\\
    \midrule
    \rowcolor{mygray}\multicolumn{5}{c}{\textbf{\textit{Mistral-7B}}}\\
        EA&72.15&7.59&-6.96&\underline{51.17}\\
         InsCL&\textbf{76.17}&\underline{4.43}&\underline{-4.02}&\textbf{54.08}\\
         RGD-R&\underline{74.91}&\textbf{4.37}&\textbf{-3.92}&50.77\\
    \midrule
    \rowcolor{mygray}\multicolumn{5}{c}{\textbf{\textit{Llama2-7B}}}\\
    EA	&76.10 &	3.52 	&-2.49 &\underline{50.91}\\
    InsCL&\underline{76.73}&\underline{2.78}&\underline{-1.96}&50.25\\
    RGD-R&\textbf{77.03}&\textbf{2.65}&\textbf{-1.25}&\textbf{51.06}\\
    \bottomrule
    \end{tabular}
    \caption{Comparison of different replay data allocation strategies. RGD-R achieves performance comparable to that of InsCL, which substantiates both its effectiveness and generalizability.}
    \label{tab:comp_with_inscl}
    \vspace{-10pt}
\end{table}

\paragraph{Comparison with Another Data Allocation Method}
The results presented in Table~\ref{tab:all_exp_res_long_seq} demonstrate the effectiveness of our proposed RGD score and RGD-R framework. To further validate the performance of RGD-R, we conduct a comparative study against the current state-of-the-art data replay method, InsCL~\citep{InsCL}, on Mistral-7B and Llama2-7B. InsCL allocates replay data based on the similarity between previous and current training tasks. As shown in Table~\ref{tab:comp_with_inscl}, RGD-R achieves comparable performance to InsCL, demonstrating the effectiveness of our proposed approach. Since our primary objective is to identify pseudo forgetting, quantify its extent through the proposed RGD metric, and try to mitigate its impact via RGD-R, the above experimental results meet our expectations. We leave exploring how to use RGD to design better continual learning algorithms for future work.

\section{Related Work}
\paragraph{Mechanism of catastrophic forgetting}

While many continual learning algorithms have been proposed, a substantial gap persists in understanding the mechanism of catastrophic forgetting. \citet{Implicit_Inference} hypothesize that models first perform ``task inference'' before applying the relevant capability, and fine-tuning biases this inference towards tasks aligned with the fine-tuning distribution, thereby suppressing performance on other prior capabilities. \citet{Interpretable_CF} believe that forgetting is primarily due to the reduced instruction-following capability, rather than a loss of task-related knowledge. Unlike our work, the above studies do not provide direct and effective evidence of pseudo forgetting on LLMs and natural language datasets.

\paragraph{Traditional methods in continual learning}
(1) \textit{Regularization-based} methods constrain the features learned from previous tasks~\citep{CPFD, IDBR} or penalize changes to weights critical for those tasks~\citep{EWC, O-LoRA}, ensuring that new learning minimally interferes with prior capability thus maintaining performance on earlier tasks. (2) \textit{Architecture-based} methods aim to reduce the interference by either increasing the model's capacity~\citep{SAPT} or isolating the existing weights~\citep{OSN}. (3) \textit{Replay-based} methods retain a small subset of prior training examples or pseudo data and revisit them when a new task is introduced~\citep{CorpusBrain, SSR, LFPT5}. InsCL~\citep{InsCL} allocates replay data based on the similarity of task instructions. In this paper, we introduce RGD-R, which dynamically allocates replay data based on the model’s susceptibility to pseudo forgetting, capturing more model-relevant
characteristics to help the model maintain both stability and plasticity.

\section{Conclusion}
In this study, we directly demonstrate the phenomenon of “pseudo forgetting” in LLMs during continual learning. We show that the performance degradation on previous tasks does not stem from the loss of corresponding capabilities, but rather from reduced instruction dependence during rationale generation. We introduce the RGD score to quantify the extent of the model’s susceptibility to pseudo forgetting, which is then used to dynamically allocate the replay ratio for each previous task to optimize replay data utilization in our proposed RGD-R framework. Experimental results confirm the effectiveness of RGD-R in addressing pseudo forgetting and preserving model plasticity.
\section*{Limitations}
While this paper analyzes and addresses pseudo forgetting during continual learning in LLMs, several limitations warrant further discussion. First, we do not conduct an in-depth analysis of the specific process behind pseudo forgetting. For instance, at what point during the learning of new tasks does the model begin to show reduced dependence on the instructions from previously learned tasks? What are the underlying factors driving this decline? Second, the relationship between pseudo forgetting and specific tasks or domains remains unexplored. For example, as noted by~\citet{words_matters}, domain generalization in summarization tasks correlates with words distribution, raising the question of whether pseudo forgetting exhibits similar characteristics. Additionally, we propose that measuring pseudo-forgetting is likely a multi-dimensional problem, and our proposed RGD score represents just one possible metric. The development of more comprehensive evaluation metrics for this phenomenon requires additional research. Finally, our findings indicate that LLMs do not forget previously acquired capabilities, and \citet{dai-etal-2022-knowledge} suggest that these capabilities are stored parametrically within the model. Consequently, to optimize continual learning algorithms, we suggest that future works could benefit from combining replay-based and parameter-based approaches, with a greater emphasis on enhancing asynchronous knowledge transfer capabilities—an underexplored aspect in current research~\citep{zhang-etal-2023-citb}.
\section*{Acknowledgments}
Supported by the Major Research Plan of the National Natural Science Foundation of China (Grant No. 92370110) and the Joint Funds of the National Natural Science Foundation of China (Grant No. U21B2009).
\section*{Ethics Statement}
This work focuses on analyzing and addressing pseudo forgetting in large language models during continual learning, and as such, does not introduce additional ethical risks beyond those inherent to standard NLP research. The potential risks primarily stem from two aspects: First, our experiments utilize large language models trained on vast amounts of internet text data, which may contain societal biases. However, since our research focuses on analyzing model capabilities rather than deploying systems, the risk of propagating harmful biases is minimal. Second, while our findings about model capabilities and instruction dependence could potentially be misused to manipulate model outputs, our work specifically aims to improve model reliability and performance stability, ultimately contributing to more robust and dependable AI systems. Throughout our experiments, we used standard benchmarks and publicly available datasets to ensure reproducibility and transparency. Our methods and findings are intended to advance the scientific understanding of continual learning in language models while adhering to established ethical guidelines in NLP research.
\bibliography{custom}
\onecolumn
\appendix
\section{Dataset Details}
\label{sec:dataset_details}
\subsection{Datasets}
\paragraph{Long Sequence Benchmark} 
The Long Sequence Benchmark~\citep{long_seq_bench} comprises 15 tasks from CL benchmark~\citep{cl_bench}, GLUE benchmark~\citep{glue_bench}, and SuperGLUE benchmark~\citep{super_glue_bench}, as detailed in Table~\ref{tab:task_detail_of_long_seq}.

\begin{table*}[h]
\centering
\begin{tabular}{lllll}
\toprule
\textbf{Dataset} & \textbf{Source} & \textbf{Task} & \textbf{Domain} & \textbf{Metric} \\
\midrule
1. Yelp & CL Benchmark & sentiment analysis & Yelp reviews & accuracy \\
2. Amazon & CL Benchmark & sentiment analysis & Amazon reviews & accuracy \\
3. DBpedia & CL Benchmark & topic classification & Wikipedia & accuracy \\
4. Yahoo & CL Benchmark & topic classification & Yahoo Q\&A & accuracy \\
5. AG News & CL Benchmark & topic classification & news & accuracy \\
6. MNLI & GLUE & natural language inference & various & accuracy \\
7. QQP & GLUE & paragraph detection & Quora & accuracy \\
8. RTE & GLUE & natural language inference & news, Wikipedia & accuracy \\
9. SST-2 & GLUE & sentiment analysis & movie reviews & accuracy \\
10. WiC & SuperGLUE & word sense disambiguation & lexical databases & accuracy \\
11. CB & SuperGLUE & natural language inference & various & accuracy \\
12. COPA & SuperGLUE & question and answering & blogs, encyclopedia & accuracy \\
13. BoolQA & SuperGLUE & boolean question and answering & Wikipedia & accuracy \\
14. MultiRC & SuperGLUE & question and answering & various & accuracy \\
15. IMDB & SuperGLUE & sentiment analysis & movie reviews & accuracy \\
\bottomrule
\end{tabular}
\caption{The details of 15 classification datasets in the Long Sequence Benchmark~\citep{long_seq_bench}.}
\label{tab:task_detail_of_long_seq}
\end{table*}

\subsection{Task Sequence Orders}
Following previous works~\citep{SAPT,long_seq_bench}, we conduct experiments using two different training orders, as shown in Table~\ref{tab:training_orders}.

\begin{table*}[h]
\centering
\begin{tabular}{cl}
\toprule
\textbf{Order} & \textbf{Task Sequence} \\
\midrule
1 & mnli $\rightarrow$ cb $\rightarrow$ wic $\rightarrow$ copa $\rightarrow$ qqp $\rightarrow$ boolqa $\rightarrow$ rte $\rightarrow$ imdb $\rightarrow$ \\
  & yelp $\rightarrow$ amazon $\rightarrow$ sst-2 $\rightarrow$ dbpedia $\rightarrow$ ag $\rightarrow$ multirc $\rightarrow$ yahoo \\
\hdashline
2 & yelp $\rightarrow$ amazon $\rightarrow$ mnli $\rightarrow$ cb $\rightarrow$ copa $\rightarrow$ qqp $\rightarrow$ rte $\rightarrow$ imdb $\rightarrow$ \\
  & sst-2 $\rightarrow$ dbpedia $\rightarrow$ ag $\rightarrow$ yahoo $\rightarrow$ multirc $\rightarrow$ boolqa $\rightarrow$ wic \\
\bottomrule
\end{tabular}
\caption{Tow different orders of task sequences used for our experiments correspond to the Long Sequence Benchmark.}
\label{tab:training_orders}
\end{table*}

\subsection{Data Construction and Ground Truth Rationales Generation}

The raw sample consists of an instruction $I$, an input $I_{input}$, and an answer $A$. We adopted the instruction conversion templates proposed by~\citet{tulu_v1} to integrate inputs into instructions ($[I,I_{input}]\to I$). To explicitly probing the model's acquired capabilities, we employed Llama3.1-70B-Instruct~\footnote{https://huggingface.co/meta-llama/Llama-3.1-70B-Instruct} to generate a rationale $R$ for each sample. The final data samples were structured as triples $(I,R_{g},A_{g})$. Specifically, we use the prompt shown in Table~\ref{fig:prompt_for_rgd_geenration} to ensure that $A_{g}$ would not appear directly within $R_{g}$, or would only appear at the end of $R_{g}$. This approach prevent the occurrence of $A_{g}$ being provided via partial rationale guidance in experiments in Section~\ref{sec:evidence_exp}, thereby ensuring the validity of our experimental results.
\begin{table*}[h]
    \centering
    \begin{tabular}{l}
\toprule
<|begin\_of\_text|><|start\_header\_id|>system<|end\_header\_id|>\\
\texttt{\{default system prompt\}}\\
<|eot\_id|><|start\_header\_id|>user<|end\_header\_id|>\\
\#\#\# Instruction:\\
\makecell[l]{We have a question and an answer provided below. Your task is to generate a rationale that explains\\the reasoning behind the given answer. The rationale should be comprehensive, logical, and clearly\\ support why the answer is appropriate for the question.}\\
\#\#\# QA Pair:\\
Original question:\\
\texttt{\{Instruction\}}\\
Original answer:\\
\texttt{\{Answer\}}\\
\#\#\# Guidelines:\\
1. Provide a detailed rationale for the given answer.\\
2. Ensure that the rationale is clear, logical, and free of any ambiguity.\\
\#\#\# Format:\\
\makecell[l]{Please generate the following JSON formatted output and nothing else:<|eot\_id|><|start\_header\_id|>\\assistant<|end\_header\_id|>}\\
\makecell[l]{\{"answer": "\texttt{\{Answer\}}", "rationale": "The correct answer is \texttt{\{Answer\}}.\\The rationale behind this answer is as follows:}\\
\bottomrule
    \end{tabular}
    \caption{The prompt for ground truth rationale generation}
    \label{fig:prompt_for_rgd_geenration}
\end{table*}

\section{Experimental Implementation Details}
\label{sec:exp_imp_detials}
\subsection{Implementation Code for the First $k$ Portions of Rationale}
\label{app:imp_k_portions}
\begin{lstlisting}[language=Python]
rationale_words = item["rationale"].split(" ")
end_part = int(len(rationale_words)*ex_rationale_rate)
part_rationale = " ".join(rationale_words[:end_part])
\end{lstlisting}
\subsection{Model Training}
To comprehensively assess the effectiveness of RGD-R, we perform comparative experiments using backbone models of different sizes and underlying knowledge bases. The backbone models used in our experiments include Qwen2-0.5B/7B~\citep{yang2024qwen2}, Mistral-7B~\citep{jiang2023mistral}, and Llama2-7B/13B~\citep{i:1-LLaMA-2}.
We perform continual learning training using the LoRa algorithm on the 7B and 13B models. Specifically, the LoRA hyperparameters are set as follows: $\texttt{lora\_rank}=8$, $\texttt{lora\_alpha}=16$, and $\texttt{lora\_dropout}=0.1$, with LoRA applied across all modules. For the Qwen2-0.5B model, we directly apply full fine-tuning. The detailed parameter settings are presented in Table~\ref{tab:training_details}:
\begin{table*}[h]
    \centering
    \begin{tabular}{lccccc}
    \toprule
    \textbf{Model Size}&\textbf{Optimizer}&\textbf{Lr Scheduler}&\textbf{Learning Rate}&\textbf{Batch Size}&\textbf{Epochs}\\
    \midrule
    $\ge7B$&AdamW&\makecell[l]{Warmup=0.03\\Decay="cosine"}&5e-4&32&6\\
    $<7B$&AdamW&\makecell[l]{Warmup=0.03\\Decay="cosine"}&5e-5&64&3\\
    \bottomrule
    \end{tabular}
    \caption{Training details of continual learning}
    \label{tab:training_details}
\end{table*}
\label{sec:Training_Details}
\subsection{GCG Implementation}
\label{sec:gcg_imp}

In Section~\ref{exp:gcg}, we employ GCG~\citep{zou2023universal} to search for the suffix corresponding to each forgotten sample, which enables the original instruction to guide the pseudo-forgetting model in generating appropriate rationale and restoring performance on previous tasks. Specifically, as shown in Table~\ref{tab:GCG_targets}, we utilize three optimization objectives to facilitate the search process. The termination conditions are set as: (1) correct model response to the original instruction, or (2) reaching the maximum iteration count of 500. We configure the hyperparameters with $\texttt{top}-k=256$ and $\texttt{batch size}=256$. The initial suffix is set to "\texttt{! ! ! ! ! ! ! ! ! ! ! ! ! ! ! ! ! ! ! ! ! ! ! }".

\begin{table*}[h]
    \centering
    \begin{tabular}{ll}
    \toprule
         \textbf{Target $T$}&  \textbf{Example ($k=0.2$)}\\
    \midrule
         Answer-based& The answer is: \texttt{\{ground truth answer\}}. The reasons are as follows:\\
         \hdashline
         Partial $R_{g}$& \makecell[l]{1. To establish the logical relationship between the two sentences, we must analyze\\the meaning and implications of each. 2. Sentence 1 states that the presence of\\ a smart doctor who gave a tip through}\\
         \hdashline
         Partial $R_{b-f}$& \makecell[l]{1. Sentence 1 states that there was a smart doctor who gave them a tip through\\the Coroner,which implies the presence and involvement of a doctor in the situation.}\\
    \bottomrule
    \end{tabular}
    \caption{Optimization targets used by GCG on MNLI task in Experiment~\ref{exp:gcg}. \textbf{1.} $R_{g}$ and $R_{b-f}$ represent the ground truth rationale and the rationale generated by the pre-forgetting model, respectively. \textbf{2.} The rationale shown here corresponds to the first 20\% of the sequence, which does not directly provide task-relevant key information.}
    \label{tab:GCG_targets}
\end{table*}
\subsection{Attribution Implementation}
\label{sec:attribution_implementation}
In Section~\ref{sec:information_loss_exp}, we quantify the model’s dependency on the given instruction during rationale generation using an attribution algorithm~\citep{rationale_drift, wang-etal-2023-label,dai-etal-2022-knowledge}. 

Specifically, we can use the Riemann approximation of the integral to calculate the contribution of a neuron $\omega$ to the model’s output $F(\cdot)$, with $m$ approximation steps:
\begin{equation}\small
\label{eq:3}
\text{Attr}(\omega)=\omega\circ\int_0^1 \frac{\partial F(\alpha \omega)}{\partial \omega} d\alpha \approx \frac{\omega}{m} \sum_{k=1}^{m} \frac{\partial F\left( \frac{k}{m} \omega \right)}{\partial \omega}
\end{equation}

Since the self-attention layers learn strong instruction-following patterns~\citep{wu-etal-2024-language}, we can compute the dependency between the instruction $I=x_1:x_{N_{ins}}$ and the given rationale $R=x_1:x_{N_{rationale}}$ based on the attention layers:
{\small
\begin{equation}
\label{eq:4}
    Q^{(l)}_{IR} = \frac{1}{|N|} \sum_{(i,j) \in D_{IR}} \text{Attr}(A^{(l)}_{i,j})\quad\text{where}\quad
D_{IR} = \{ (i,j) | x_i\in I, x_j \in R\}
\end{equation}
}

In this notation, $\text{Attr}(A^{(l)}_{i,j})$ represents the dependence intensity from the $i$-th token to the $j$-th token in the $l$-th self-attention layer, calculated by summing the absolute attribution scores across all heads. $|N|$ denotes the total number of rationale steps.

In Equation (~\ref{eq:3}), $F(\cdot)$represents the language modeling loss, and $m=20$. Each sentence in the rationale is treated as a separate reasoning step, allowing us to compute the total number of inference steps, $|N|$, as described in Equation (\ref{eq:4}).
\section{Additional Experiments}
\subsection{Evaluation of Task Information Provided by the First $k$ Portions of Rationale}
\label{app:eval_k_rationale}
\begin{figure}[th]
    \centering
    \includegraphics[width=1\linewidth]{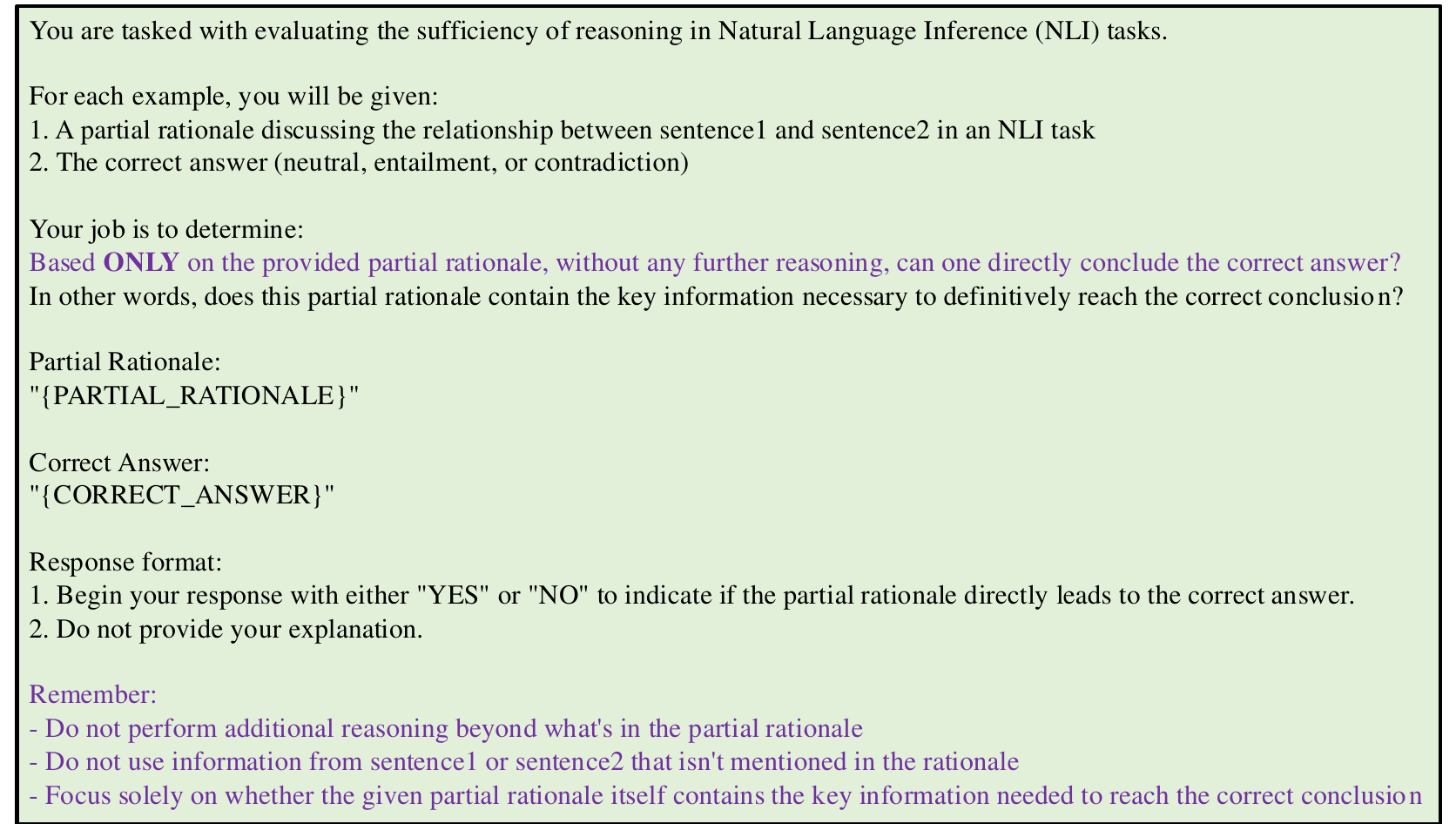}
    \caption{The prompt used to assess whether the first $k$ portions of rationales directly provide task-related key information. We ask GPT4o to evaluate whether the correct answer can be directly obtained based solely on the provided rationale, without requiring further reasoning.}
    \label{fig:promtp_for_information_eval}
\end{figure}

In experiment A1 described in Section~\ref{exp:A1}, as illustrated in Figure~\ref{fig:ex_knowledge_exp_res_llama}, when the forgetting model is provided with the first $k$ portions of the rationale, its performance on the forgotten tasks gradually recovers to pre-forgetting levels as $k$ increases. However, since the first $k$ portions of the rationale may introduce task-relevant critical information, the results from experiment A1 cannot directly prove the existence of pseudo forgetting. Nevertheless, A1 motivates us to conduct the A2 experiment, enabling the model to actively generate appropriate rationale and derive the correct answers by searching for task-irrelevant suffixes. 

To ensure the validity of the A2 experiment, we evaluate the proportion of external rationales that do not introduce key information for various values of $k$ using GPT4o~\footnote{https://openai.com/index/gpt-4o-system-card/}. Subsequently, we perform GCG using the $k$ value that does not leak any information, ensuring that the suffixes do not encode any task-critical information but serve merely to guide the model’s rationale generation.

\paragraph{Experimental setup} We randomly sample 100 ground truth rationales from both the MNLI and RTE respectively. We use GPT4o to assess whether the first $k$ portions of rationales provide sufficient critical information to obtain the correct answer without further reasoning. (The prompt is shown in Figure~\ref{fig:promtp_for_information_eval})

\begin{table}[h]
    \centering
    \begin{tabular}{lcccccc}
    \toprule
        \textbf{Task} &  $k=0.1$ &  $k=0.2$	& $k=0.3$	& $k=0.4$	& $k=0.5$	& $k=0.6$\\
    \midrule
        \textbf{MNLI} &96.0	&94.0&84.0&73.0&49.0&24.0 \\
        \textbf{RTE} &97.0&97.0&87.0&78.0&57.0&36.0  \\
        \textbf{AVG} &96.5&95.5&85.5&75.5&53.0&30.0  \\
    \bottomrule
    \end{tabular}
    \caption{Percentage (\%) of cases where the first $k$ portions of rationales do \textbf{NOT} provide critical information. When  $k = 0.1$  or  $k = 0.2$, the key information leakage rate is around 5\%, which is acceptable. Therefore, in Experiment A2 in Section~\ref{exp:gcg}, we use the first 0.2 portions of the rationale as the optimization target for GCG, examples are shown in Table~\ref{tab:GCG_targets}.}
    \label{tab:evaluation_of_k_rationale}
\end{table}

\paragraph{Experimental results and analysis} 
The experimental results are shown in Table~\ref{tab:evaluation_of_k_rationale}. When $k\leq0.4$, the first $k$ portions of rationales generally do not directly provide task-relevant critical information. When $k\ge0.5$, more than half of the first $k$ portions contain some task-critical information, which aligns with intuition. To ensure that no external key information is introduced, we set $k=0.2$  in Experiment A2, using the first 20\% words of the rationale as the optimization target for GCG, and search for meaningless instruction suffixes.
\subsection{More Results for Attribution Experiments}
In Section~\ref{sec:information_loss_exp}, we employ an attribution algorithm to investigate how much the model relies on task instructions during the rationale generation stage, both before and after pseudo forgetting. Our findings reveal a significant decline in instruction dependency for pseudo-forgetting models, which in turn impairs the model’s ability to correctly utilize relevant task-specific abilities when prompted with the original instructions. This degradation contributes directly to the observed pseudo forgetting phenomenon.

\begin{figure}[t]
    \centering
    \begin{minipage}[t]{0.48\textwidth}
        \centering
        \subfigure[Llama2-7B]{\includegraphics[width=0.48\textwidth]{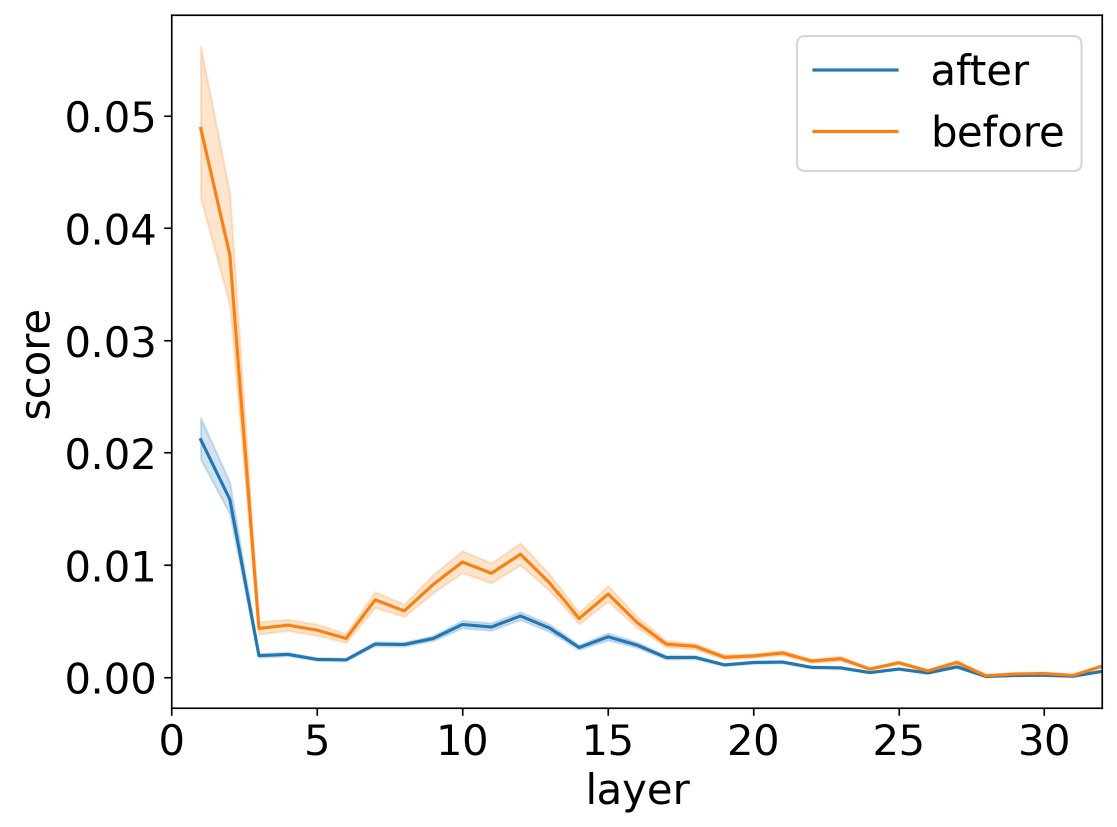}}
        \subfigure[Mistral-7B]{\includegraphics[width=0.49\textwidth]{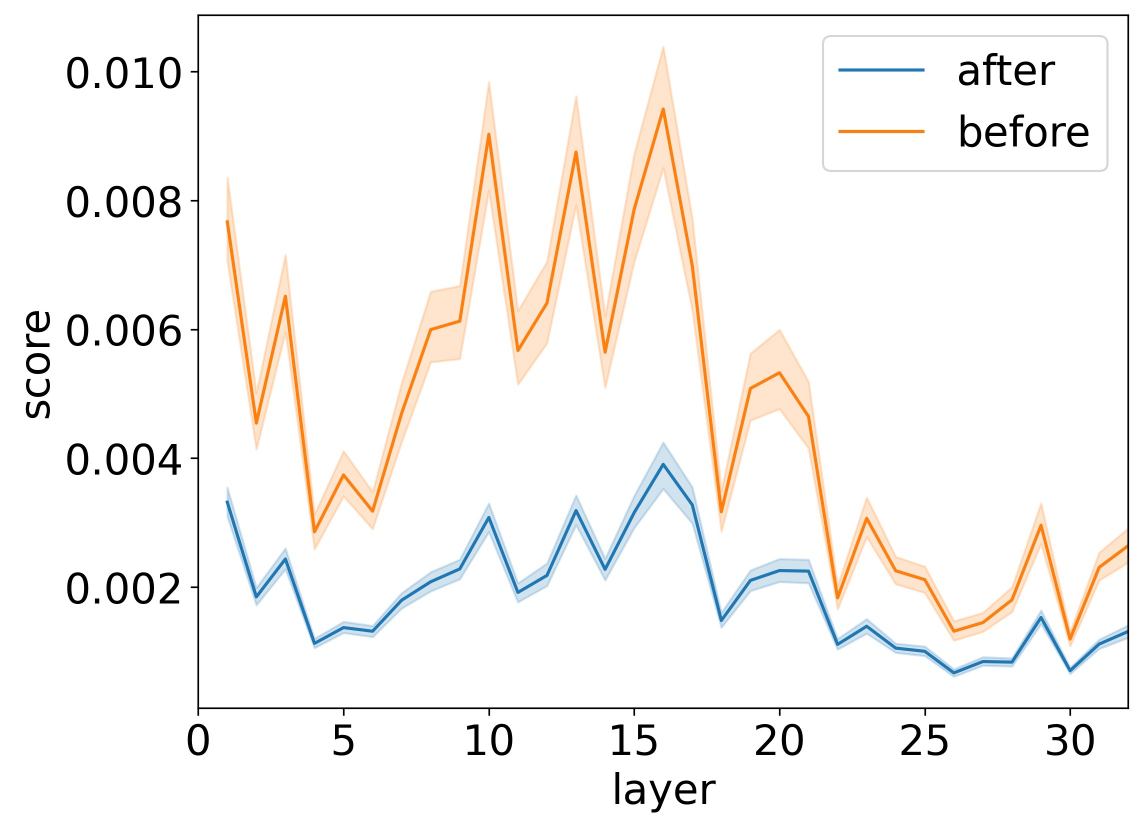}}
        \caption{Comparison of instruction dependency scores of pseudo-forgetting model for generating correct and incorrect rationales on RTE task.}
        \label{fig:more_pred_attri_on_rte}
    \end{minipage}
    \hfill
    \begin{minipage}[t]{0.48\textwidth}
        \centering
        \subfigure[Llama2-7B]{\includegraphics[width=0.48\textwidth]{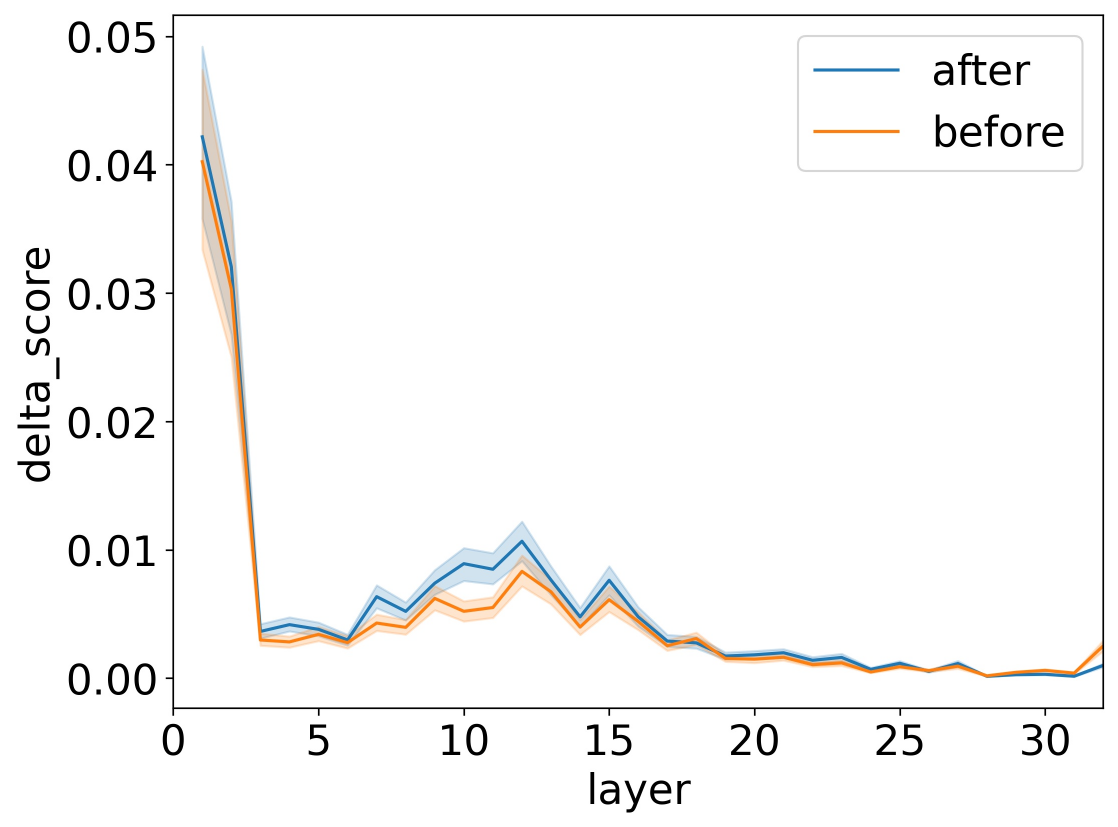}}
        \subfigure[Mistral-7B]{\includegraphics[width=0.49\textwidth]{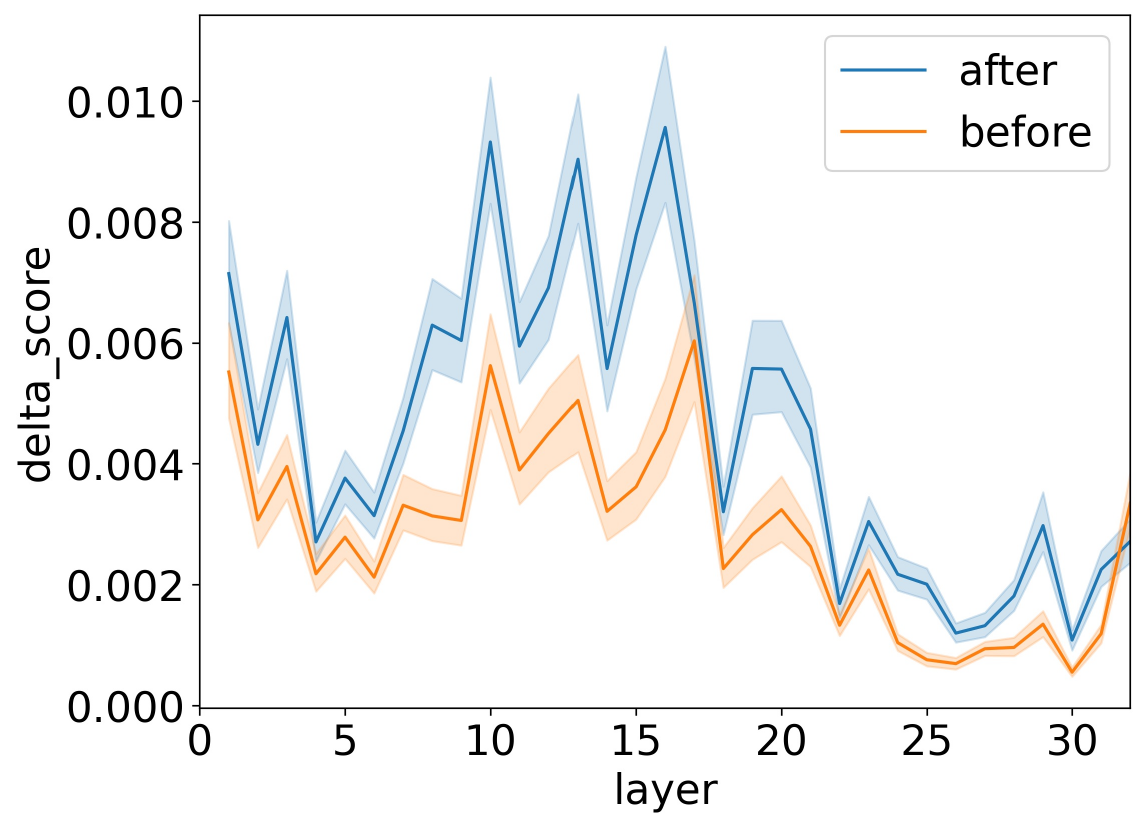}}
        \caption{Comparison of relative instruction dependency scores across different states of Llama2-7B and Mistral-7B on RTE task.}
        \label{fig:more_pred_delta_attri_on_rte}
    \end{minipage}
\end{figure}

Figure~\ref{fig:more_pred_attri_on_rte} and Figure~\ref{fig:more_pred_delta_attri_on_rte} present the results of Experiment 1 and Experiment 2 in Section~\ref{sec:information_loss_exp} on the RTE task, respectively. The observed trends are consistent with those in Figure~\ref{fig:pre_scre_res} and Figure~\ref{fig:delta_scre_res}, similarly supporting our findings.

\section{Case study}
\label{sec:case_study}

\begin{table*}[h]
    \centering
    \begin{tabular}{ccl}
    \toprule
    \bf Model&\bf Task&\bf Partial suffixes\\
    \midrule
         Mistral-7B&BoolQA & \texttt{! ! Sounds striking ! ! ! ! ! Bo ..\#\# !phony provisions !="\#}\\
         Mistral-7B&BoolQA & \texttt{And ! ! ! ! ! doesn ! mentioned ! !However ! ! ! Shadow ! ! }
         \\
         Mistral-7B&MNLI &  \texttt{! ! ! ! ! ! ! ! ! the ! ! Fifth ! ! ! ! ! ! !}
         \\
         Mistral-7B&MNLI&  \texttt{! ! Cons ! > nation ! April ! G ! Pub Final ! ! ! ! ! ! !}
         \\
         Qwen2-0.5B&MNLI&\texttt{!HolAndHashCode ! ErrorResponse-not Donovan unpublished }
         \\
    \bottomrule
    \end{tabular}
    \caption{Examples of instruction suffixes discovered by GCG. Due to length constraints, only the initial portions of the suffixes are shown.}
    \label{tab:Explored_suffix_cases}
\end{table*}
\end{document}